\documentclass{article}

\usepackage{microtype}
\usepackage{graphicx}
\usepackage{subfigure}
\usepackage{booktabs} 
\usepackage{amsmath}
\usepackage{amssymb}
\usepackage{amsthm}

\usepackage{amsmath}

\usepackage{hyperref}

\DeclareMathOperator*{\expect}{\mathbb{E}}

\DeclareMathOperator*{\ent}{\text{ent}}

\usepackage[accepted]{icml2018}

\icmltitlerunning{A Deep Generative Model for Functions}

\begin{document}

\twocolumn[
\icmltitle{VFunc: A Deep Generative Model for Functions}


\begin{icmlauthorlist}
\icmlauthor{Philip Bachman}{msr}
\icmlauthor{Riashat Islam}{mcgill,mila}
\icmlauthor{Alessandro Sordoni}{msr}
\icmlauthor{Zafarali Ahmed}{mcgill,mila}
\end{icmlauthorlist}
\icmlaffiliation{mila}{Montreal Institute of Learning Algorithms}
\icmlaffiliation{mcgill}{School of Computer Science, McGill University}
\icmlaffiliation{msr}{Microsoft Research}
\icmlcorrespondingauthor{Philip Bachman}{phil.bachman@gmail.com}


\vskip 0.3in
]



\printAffiliationsAndNotice{}  

\begin{abstract}
We introduce a deep generative model for functions.
Our model provides a joint distribution $p(f, z)$ over functions $f$ and latent variables $z$ which lets us efficiently sample from the marginal $p(f)$ and maximize a variational lower bound on the entropy $H(f)$.
We can thus maximize objectives of the form $\expect_{f \sim p(f)}[R(f)] + \lambda H(f)$, where $R(f)$ denotes, e.g., a data log-likelihood term or an expected reward.
Such objectives encompass Bayesian deep learning in function space, rather than parameter space, and Bayesian deep RL with representations of uncertainty that offer benefits over bootstrapping and parameter noise. In this short paper we describe our model, situate it in the context of prior work, and present proof-of-concept experiments for regression and RL.
\end{abstract}

\section{Introduction}
Forming better estimates of predictive uncertainty for deep neural networks has been the focus of much recent work~\cite{blundell2015,gal2016dropout,louizos2017multiplicative}. Models capable of representing confidence in their own predictions are useful in many cases, such as for avoiding over-fitting~\cite{blundell2015}, for guiding strategies for active learning~\cite{gal2017deep}, and for enhancing exploration in reinforcement learning (RL)~\cite{osband2016deep}. Better confidence estimates in powerful models would also lead to safer systems which fall back to a human operator when prediction confidence is low~\cite{xie2007predicting,kendall2017uncertainties}.

Bayesian neural networks~\cite{neal2012bayesian} estimate uncertainty by placing a prior distribution over the network parameters and then attempting to model a posterior over the parameters given some data. Due to the large number of parameters and their non-trivial correlations in the true posterior, exact inference is generally intractable and approximate inference techniques must be used. These include Markov Chain Monte Carlo with Hamiltonian Dynamics~\cite{neal2012bayesian} and variational inference ~\cite{blundell2015,gal2016dropout,pawlowski2017implicit}. Mean field approximations are often employed due to the challenge of constructing flexible approximate posteriors~\cite{blundell2015}. Neural networks are typically over-parametrized, so different parameters yield the same prediction function~\cite{dinh2017sharp}, which may produce highly multi-modal posteriors.

We argue that capturing uncertainty directly over functions, rather than over parameters, may be advantageous. E.g., a particular function may be induced by many sets of parameters, while each set of parameters induces only one function. In this sense, it is more parsimonious to model uncertainty in function space than in parameter space.
To develop intuition about what it means to model distributions over functions, what priors over functions might look, and why they are useful for, e.g.~RL, we recommended readers acquaint themselves with Gaussian Processes \cite{Bishop06} and Posterior Sampling \cite{Russo17}.

Rather than approximating a posterior over model parameters, we propose maintaining a posterior over functions\footnote{We use \emph{function} to denote a particular mapping from each possible input to a single output. Different inputs may map to different outputs, and outputs may be multi-dimensional.}, modeled by deep neural networks. Samples from our approximate posterior are obtained by first drawing a set of latent variables from a simple prior distribution and then sampling a function conditioned on those variables. Sampling a function for a single input corresponds to conditioning the computation for that input on the latent variables and sampling from the corresponding output distribution. When the latent variables are held fixed and this sampling is repeated across the whole input domain, producing a single sampled output for each possible input, we are sampling a single function from the space of input$\rightarrow$output mappings represented by our model.

We use tools from stochastic gradient variational inference~\cite{kingma2013auto} to maximize lower bounds on the entropy of the approximate posterior~\cite{barber2003algorithm}. We apply our model in regression and RL settings. In regression, we maximize variational lower bounds on data log-likelihood. In RL, we learn a posterior over policies, which could be used to perform Thompson sampling~\cite{thompson1933likelihood,fortunato2017, plappert2017parameter,riquelme2018deep}. We present proof-of-concept experiments showing that the learned latent variables encode different policies, each corresponding to a successful strategy for achieving high reward.

\section{Related Work}
We represent distributions over functions by modelling a joint distribution $p(f, z) = p(f|z) p(z)$ over functions $f$ and latent variables $z$. This provides a model-based approach similar to Gaussian Processes, which define a prior over functions which permits (sometimes) tractable computation of a full posterior \cite{gaussianprocesses}. Gaussian Processes are a non-parametric kernel-based method, which have notable benefits, but also weaknesses due to dependence on fixed data representations and scaling problems. In contrast, we sample latent variables from a prior distribution and use them to condition the computation of a neural network model $p(y | x, z)$ which puts a distribution over outputs given an input and latent variable. In contrast with most prior work on Bayesian neural networks \cite{blundell2015, gal2016dropout, BHNs}, we model approximate posteriors over functions rather than model parameters. We capture prediction uncertainty through the joint effects of entropy in the latent prior $p(z)$ and the conditional prediction $p(y | x, z)$. At a given $x$, the predictive distribution and uncertainty can be obtained by marginalizing over $z$.

Our motivation and model are closely related to recent work on latent-conditioned policies \cite{hausman2018learning,florensa2017stochastic,DIAYN,latentpolicies,MAESN, gupta2018meta}, and to similar work on maximizing entropy-regularized objectives in RL \cite{nachum2017bridging,neu2017unified,SAC}. We model distributions over mappings from the full set of states to a corresponding set of action probabilities\footnote{We can consider functions mapping states to action logits, to action probabilities, or to actual actions. The ``output space'' of functions generated by our model is a design choice.}, which differs slightly from the MaxEnt-motivated works cited here, which model distributions over mappings from individual states to action probabilities. Our approach differs from work on maximizing mutual information between latent variables and state visitation patterns in several ways. E.g., our entropy maximization objective does not require RL-type optimization, since we encourage diversity among policies in terms of state$\rightarrow$action mapping rather than how these mappings affect state visitation patterns. Our objective is thus easier to train but may lack benefits of explicitly encouraging dispersion over the state space.

Concurrent work in \cite{garnelo2018} proposes a model with the same basic components as ours. However, they focus on training $p(f)$ to maximize log-likelihood for some observed functions -- analogous to training a standard VAE to model a distribution of images. Adapting our model to the problems they consider is straightforward.

\section{Method Description}
Our model relies on a \emph{prediction network} $p(y | x, z)$ which stochastically maps inputs $x$ to outputs $y$ conditioned on latent variables $z$, a prior distribution $p(z)$, a \emph{recognition network} $q(z | f)$ which encodes a sampled function $f$ and predicts the $z$ which generated $f$, and a prior distribution $\bar{p}(f)$ over the relevant function space.
These components combine to approximate the true posterior over functions $p^{\ast}(f|D)$, given data $D$, by optimizing a variational lower bound, similar to~\citep{blundell2015}, $\mathcal{F} = \mathbb{E}_{f \sim p(f)} [ R(D|f) + \log \bar{p}(f) ] + H(f)$, comprising a data-dependent likelihood term (or expected reward in the case of RL tasks), the prior log-likelihood and the posterior entropy.
Optimizing the data-dependent term is straight-forward, e.g.~we can use standard maximum likelihood or policy gradient.
For our bound, $\bar{p}(f)$ can be an unnormalized energy. I.e., we can use arbitrary regularizers which assign lower values to ``good'' functions and higher values to ``bad'' functions. E.g., we can use norms of Jacobians of the input$\rightarrow$output mapping or model-specific qualities like robustness to perturbation \cite{Bachman2014}. We now describe a lower bound for the entropy term $H(f)$.

\subsection{Variational Lower Bound on Entropy}
We first write the mutual information $I(f; z)$:
\begin{equation}
I(f; z) = H(f) - H(f|z) = H(z) - H(z|f),
\end{equation}
and shuffle terms to get the marginal entropy of $f$:
\begin{equation}
H(f) = H(z) - H(z|f) + H(f|z).
\label{eq:entropy_of_f}
\end{equation}
For now, we consider $H(z)$ fixed -- it is entropy of the prior, and we design our model so $H(f|z)$ is easy to compute and optimize. E.g., we can make $p(y|x,z)$ a univariate Gaussian, which makes $H(f|z)$ a simple sum of independent Gaussian entropies\footnote{We consider sums and ignore integration issues. We think this is reasonable, since all digital computation is over discrete sets.}. The trickier term in Eqn.~\ref{eq:entropy_of_f} is $-H(z|f)$. Since we want a lower bound on $H(f)$, we need an upper bound on $H(z|f)$. We get such a bound by replacing conditional entropy with conditional cross entropy:
\begin{align}
H(z|f) &= \expect_{(f, z) \sim p(f,z)} [-\log p(z|f)] \\
\, &\leq \expect_{(f, z) \sim p(f,z)} [-\log q(z|f)],
\label{eq:xent_bound_on_ent}
\end{align}
where $p(z|f)$ is the true posterior over $z$ given a function $f$ in the joint $p(f,z)$ generated by our model and $q(z|f)$ is provided by the trainable recognition network. Combining Eqn.~\ref{eq:xent_bound_on_ent} with Eqn.~\ref{eq:entropy_of_f}, we get a lower bound on $H(f)$:
\begin{equation}
H(f) \geq H(z) + \expect_{(f, z) \sim p(f,z)} [\log q(z|f)] + H(f|z)
\label{eq:variational_ent_bound}
\end{equation}
The main challenges for our model are (i) how to generate functions given a latent variable and (ii) how to encode a sampled function so that we can predict the $z$ which generated it. We now describe our approach to (i) and (ii).

\subsection{Generating a Function}
We sample a function by sampling a $z$ from $p(z)$ and then conditioning the computation for any $x$ on $z$ using the prediction network $p(y|x,z)$. For optimizing the bound in Eqn.~\ref{eq:variational_ent_bound}, we sample $z$ from $p(z)$, then sample a set of $k$ points $\{x_1,...,x_k\}$ from the input domain, and then sample a $y_n \sim p(y|x_n,z)$ for each $x_n$ to get a \emph{partially observed function} $\hat{f} = \{(x_1, y_1),...,(x_k,y_k)\}$. A partially observed function $\hat{f}$ can be provided as input to the recognition network, i.e.~we can use $q(z|\hat{f})$ in place of $q(z|f)$. Using partial functions rather than full functions obtains tractability in exchange for a looser bound.

\subsection{Encoding a Function}
To predict the $z$ which generated a partially observed function $\hat{f}$, we encode the function and then predict $z$ given that encoding. To encode $\hat{f} = \{(x_1,y_1),...,(x_k,y_k)\}$, we first condition the prediction network on a \emph{default} latent variable $\bar{z}$, then backpropagate through a loss $L(y_n, p(y|x_n,\bar{z}))$ which compares the default prediction for $x_n$ to the value $y_n$ observed in $\hat{f}$ to get $\frac{\partial L_n}{\partial \bar{x}}$, and finally sum over $x_n \in \hat{f}$ to get a ``diff gradient'' which describes the difference between $\hat{f}$ and the default function in terms of gradients on the default latent variable $\bar{z}$. We then pass the diff gradient through an additional MLP to get the final $z$ reconstruction $q(z|\hat{f})$. The generality of encoding \emph{sets} of input/output pairs via backpropagation through the network that generated them can be supported by reference to work showing the universality of gradient-based metalearning \cite{Finn2018} and the universality of bag-of-words-based representations for functions with set-valued inputs \cite{Zaheer2017}.

\subsection{Dynamic Discretization Bound}
\label{sec:dd_bound_main_text}
By repeatedly sampling sets of $z$s from the prior, then sampling a single $\hat{f}$ given one of the $z$s, and then trying to pick out which $z$ in the set generated $\hat{f}$, we can maximize an expectation over lower bounds on mutual information in subsets of the full joint, which then provides a lower bound on $I(f;z)$. This \emph{dynamic discretization} bound is similar to Mutual Information Neural Estimation \cite{Belghazi2018}, but derived as approximate variational inference. We provide details in the appendix (see: Sec.~\ref{sec:dd_bound_appendix}).

\section{Experiments}
We first examine our model on some simple regression tasks, and then in the RL context on several gridworld tasks. Our model successfully maintains diverse posteriors over functions and policies which perform well on the tasks.

\subsection{Regression Tasks}
We evaluate how well our model captures uncertainty on a toy regression task from~\cite{blundell2015}. In this setting, $p(y|x,z)$ takes a scalar as input and outputs the mean and log-variance of a univariate Gaussian. Our model can reasonably capture the uncertainty and trend of the real function compared to a few baselines: Bayes by Backprop~\cite{blundell2015}, MCDropout~\cite{gal2016dropout} and Bayes by Hypernet~\cite{pawlowski2017implicit}~\footnote{Baseline implementations taken from~\cite{pawlowski2017implicit}}.

\begin{figure}
    \centering
    \includegraphics[scale=0.22]{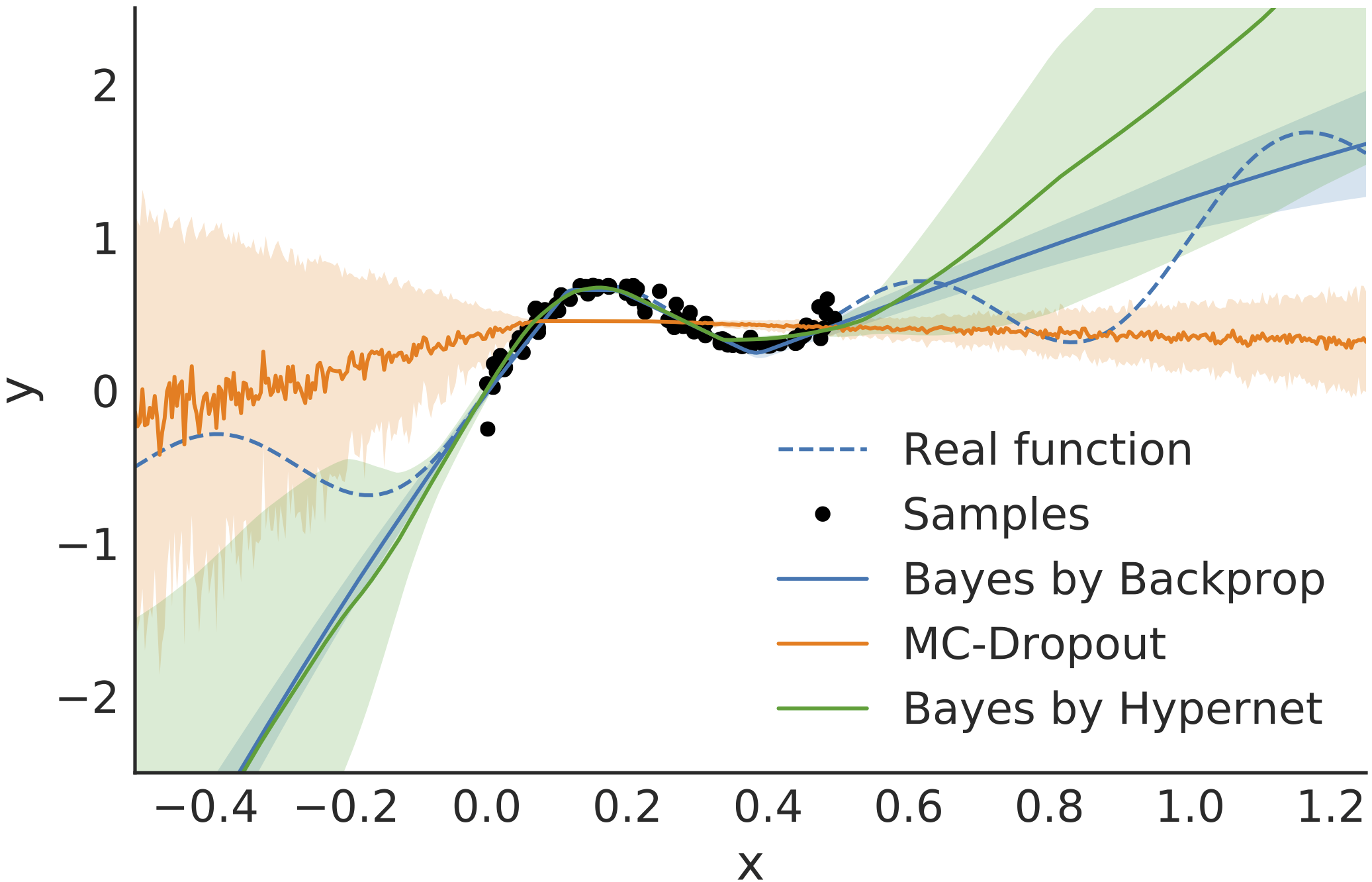}
    \includegraphics[scale=0.25]{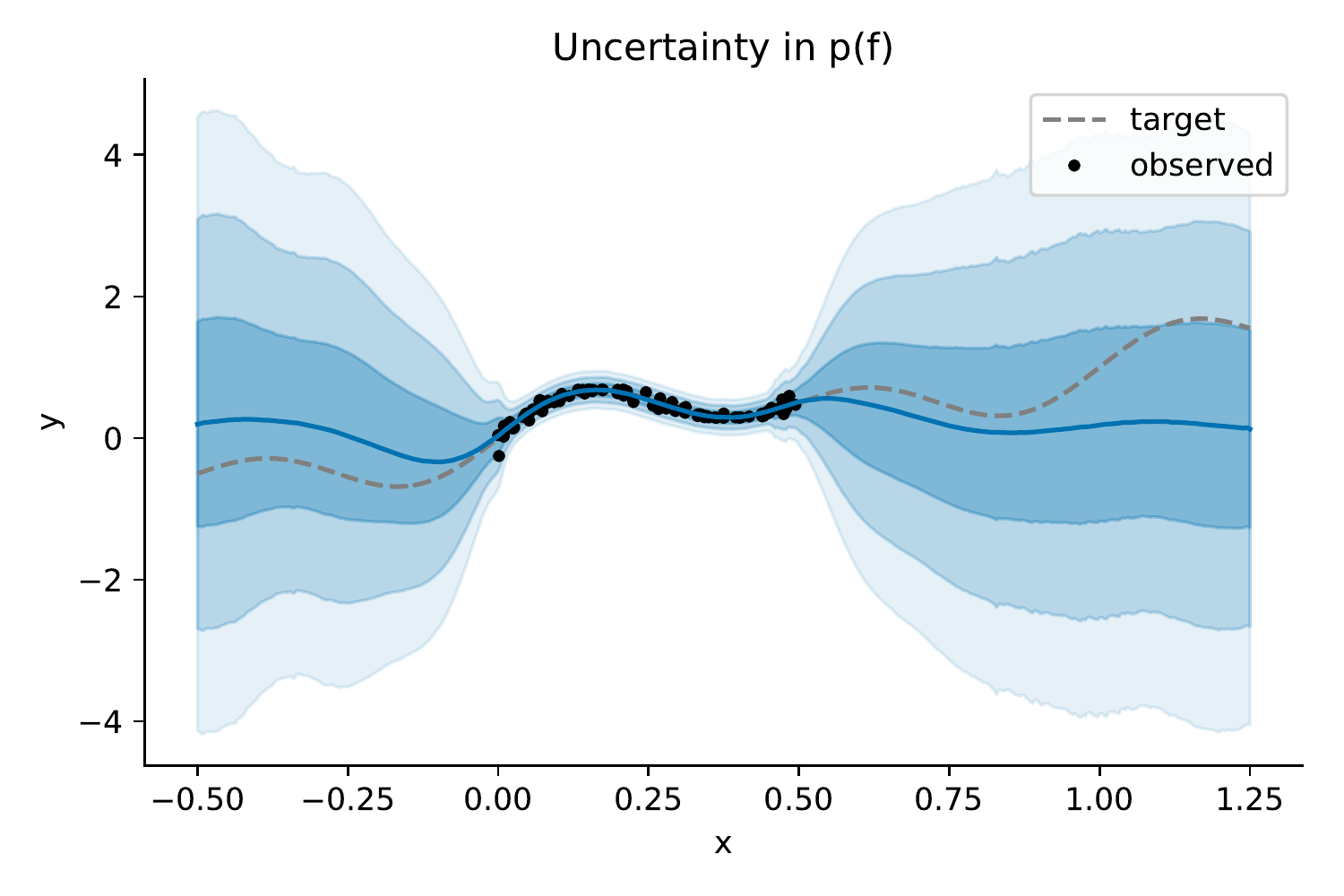}
    \includegraphics[scale=0.35]{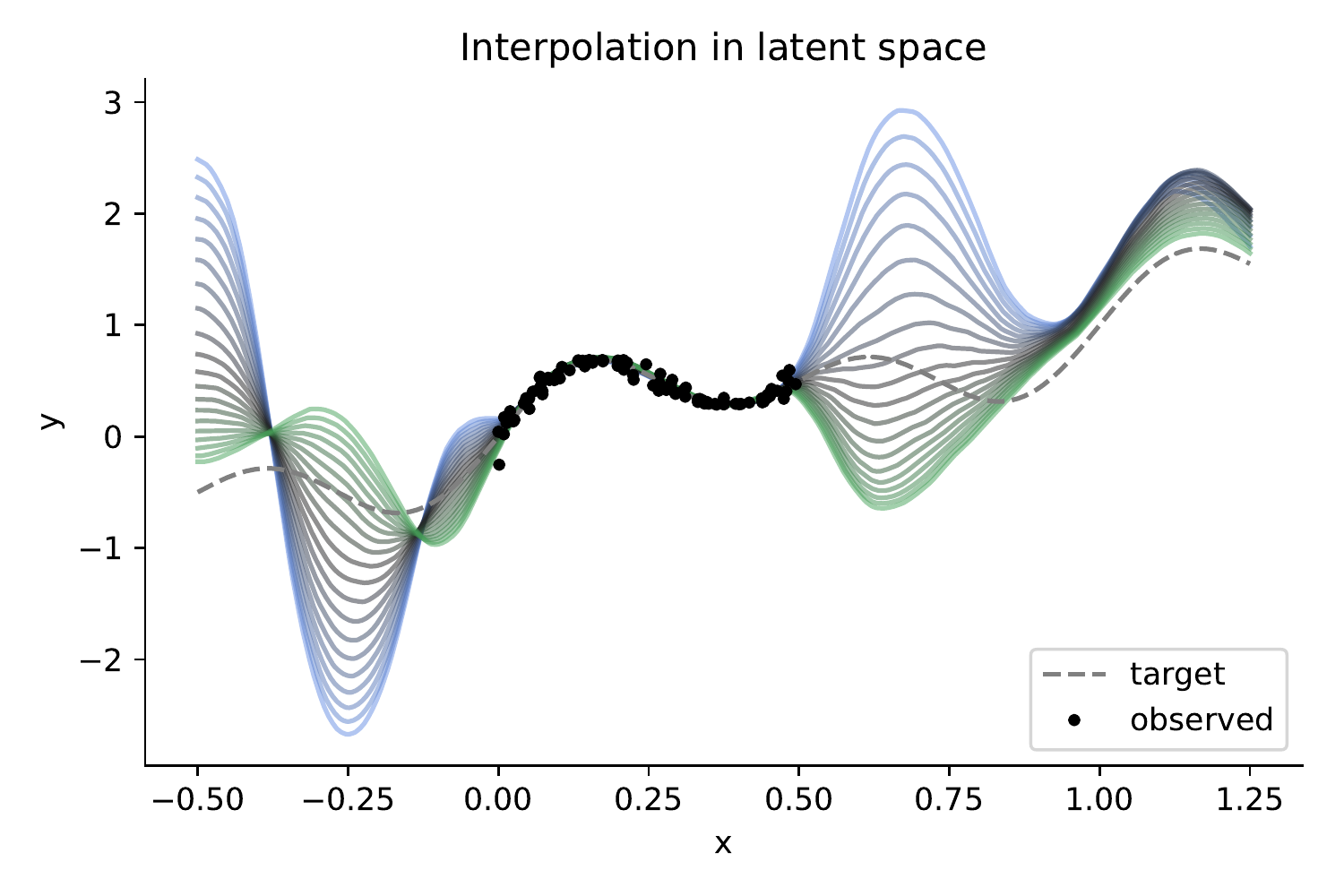}
    \caption{Predictive uncertainty captured by the baselines \textbf{(top, left)} and our model \textbf{(top, right)} for the regression task from~\cite{blundell2015}. Shaded regions correspond to standard deviations. \textbf{(bottom)} Interpolation in the latent space of functions induced by our model. We plot the mean of the predictive Gaussian $p(y | x, z)$, for $z$ values obtained by interpolating between $z_1, z_2 \sim p(z)$.}
    \label{fig:blundell_data}
\end{figure}

To visualize how the latent variables affect the generated functions, we interpolate between pairs of latent variables sampled from $p(z)$ and plot the functions induced by the interpolated $z$s. Figure ~\ref{fig:blundell_data} (bottom) shows that $z$ captures more uncertainty in regions with less training data.

\subsection{Reinforcement Learning Tasks}
\begin{figure*}[htp]
    \centering
    \includegraphics[scale=0.45]{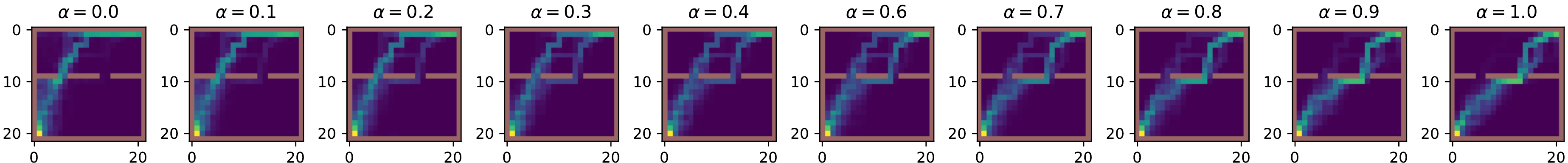}
    \caption{Interpolating in the latent space of policies in the double-slit world. When $\alpha=0$ policies tend to go through one of the slits. As $\alpha$ increases, policies gradually move toward the other slit with a nice split around $\alpha=0.5$. We plot state visitation frequencies.}
    \label{fig:RL_maxent_interpolation}
    \vspace{-0.5cm}
\end{figure*}

For RL tasks, our model represents a distribution over policies $p(\pi, z)$. Our prediction network is a latent-conditioned policy $\pi(a | s, z)$ which maps each state to a distribution over actions.
Similarly to the regression setting, our optimization objective includes a reward term and an entropy term, $H(p(\pi))$.
We maximize the expected reward for policies $\pi \sim p(\pi)$ using simple policy gradient methods~\cite{williams1992simple}.
To maximize entropy $H(p(\pi))$, we sample policies from the posterior and compute their action probabilities on states sampled from a replay buffer.
We use the resulting pairs of states and action probabilities as partially observed functions $\hat{f}$ for maximizing the lower bound on entropy $H(p(\pi))$ as detailed in Section 3.3.

\begin{figure}
    \centering
    \footnotesize{(a) $\lambda_{\ent}=0$~~~~~~~~~~~~~~~~~~~~~~~~~(b) $\lambda_{\ent}=1$}\\
    \includegraphics[scale=0.16]{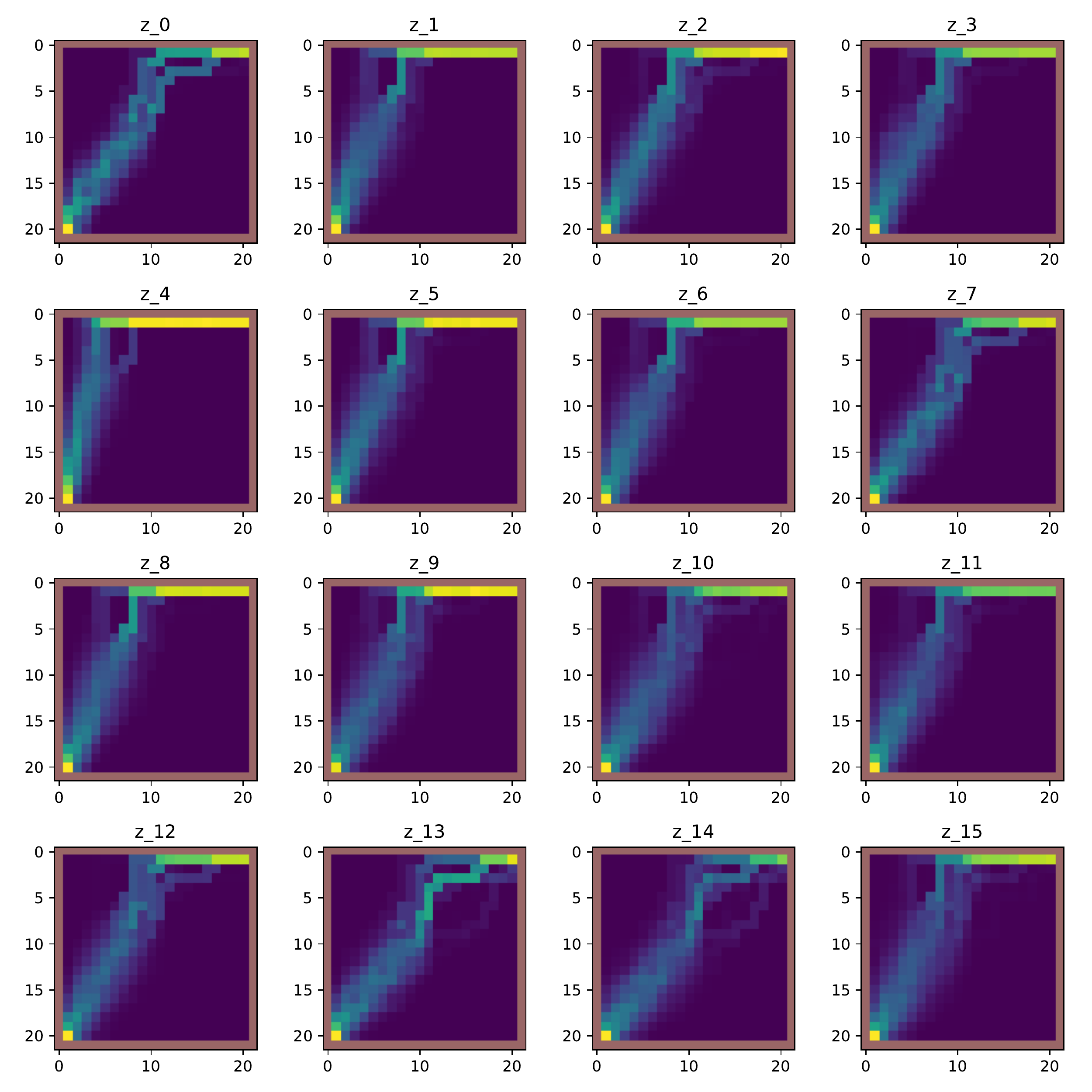}
    \includegraphics[scale=0.16]{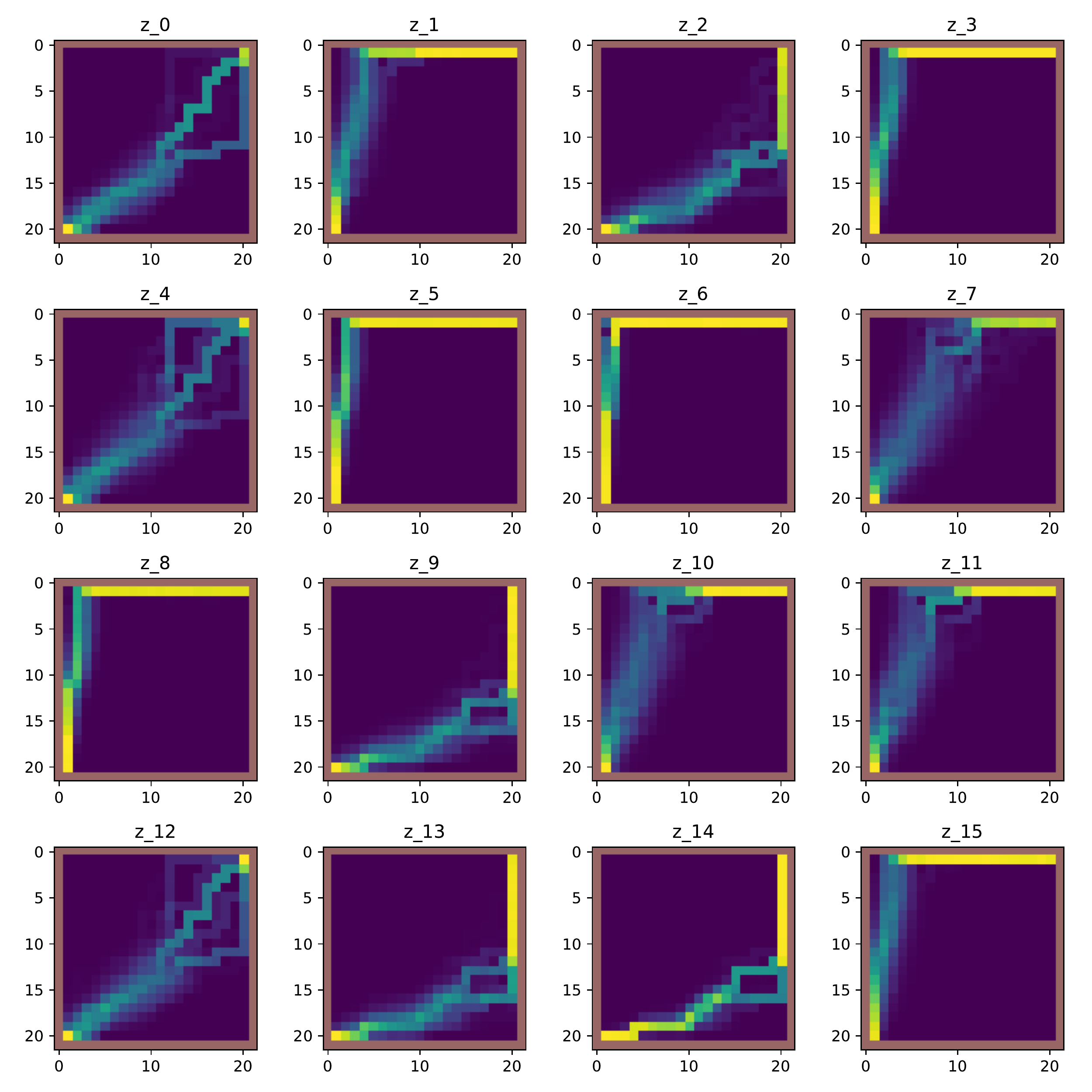}
    \vspace{-3mm}
    \caption{16 policies sampled from $p(\pi, z) = p(\pi|z)p(z)$ on an empty gridworld, (a) without entropy maximization and (b) with entropy maximization, which ensures that generated policies are more diverse. We plot state visitation frequencies.}
    \label{fig:RL_SimpleWorlds}
    \vspace{-0.5cm}
\end{figure}

\begin{figure}
    \centering
    \includegraphics[scale=0.16]{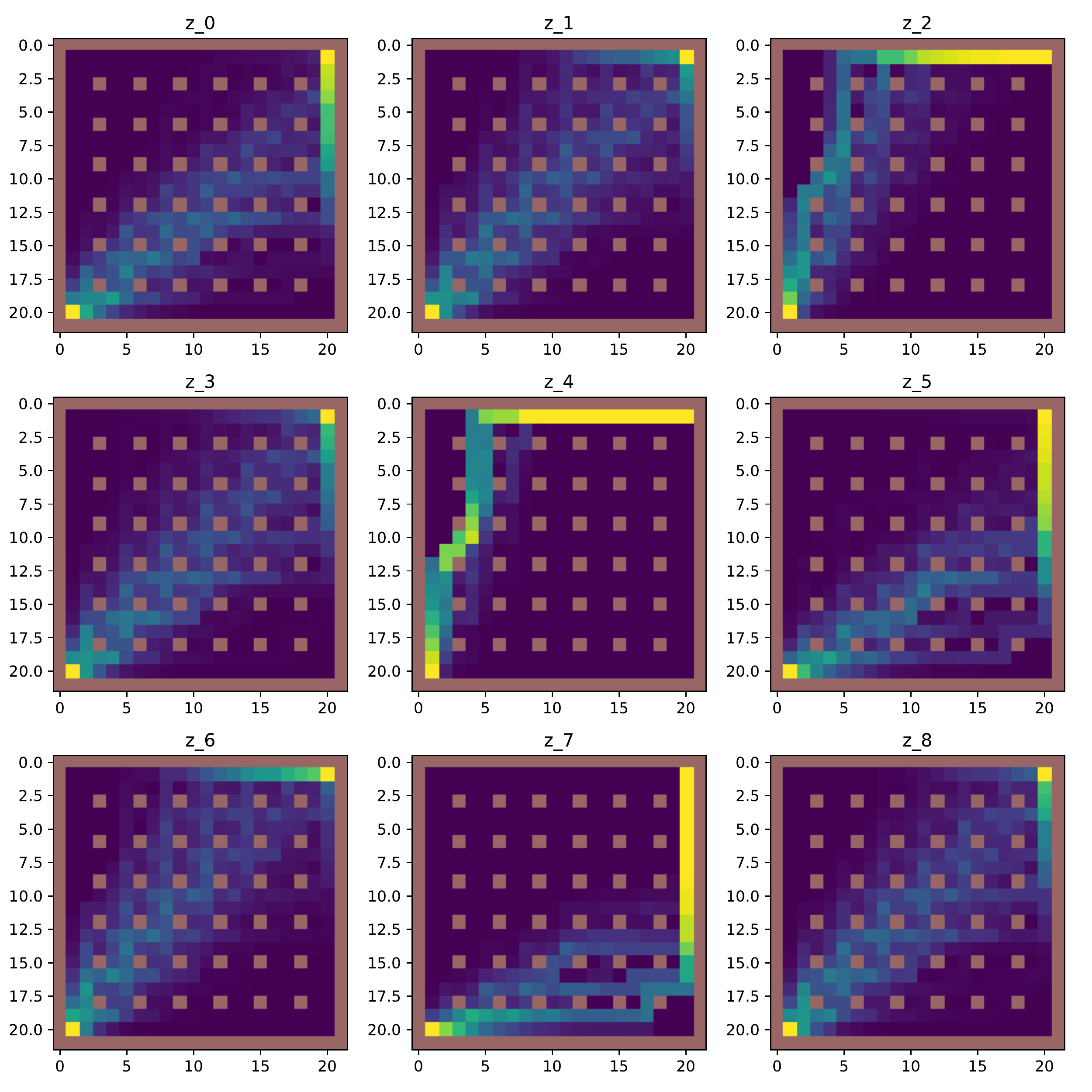}
    \includegraphics[scale=0.16]{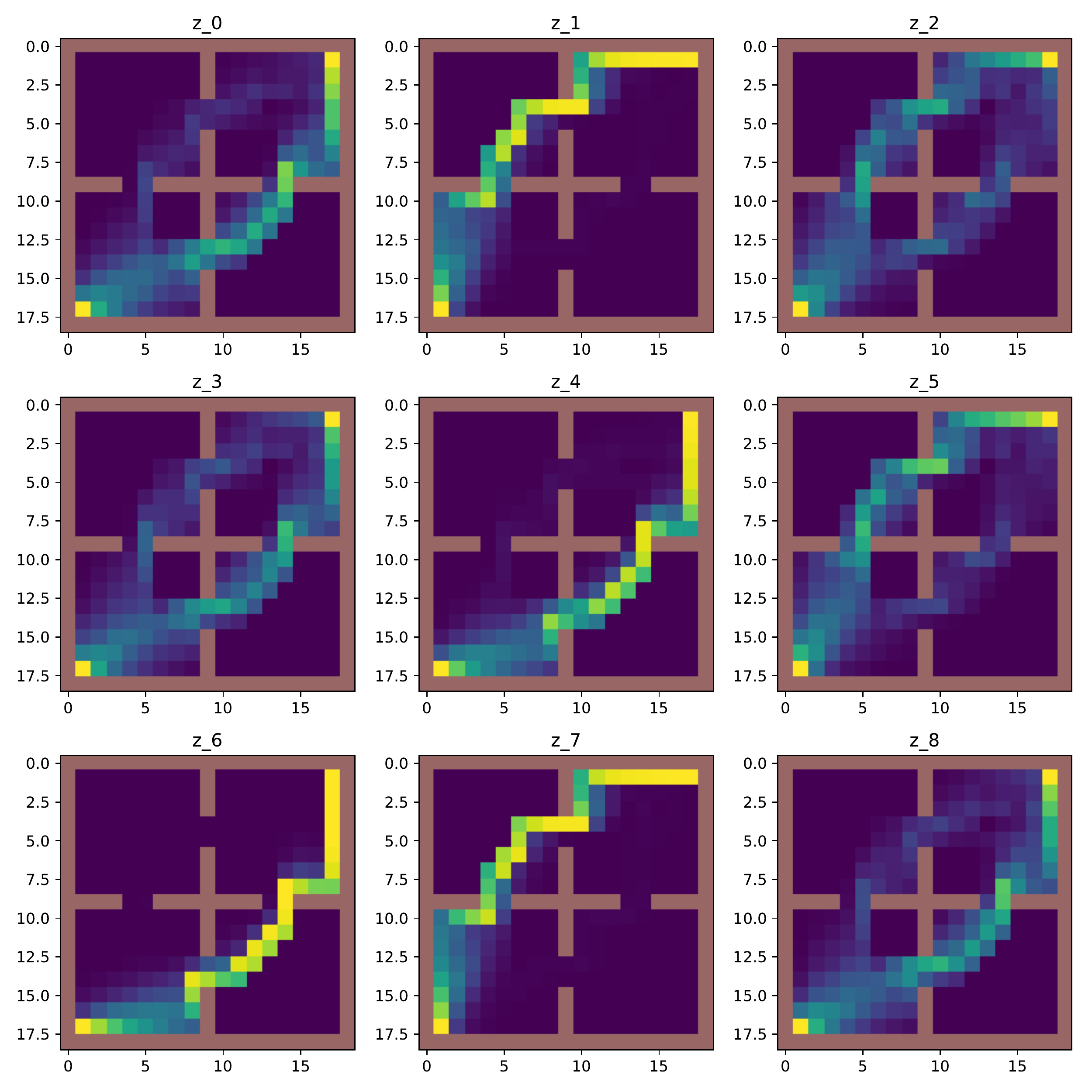}
    \vspace{-5mm}
    \caption{State visitation frequencies given nine $z$s sampled independently from $p(z)$ for the pachinko and four rooms environment.}
    \vspace{-0.5cm}
    \label{fig:RL_manywalls}
\end{figure}

Our experimental setting comprises various gridworld domains implemented with Easy MDP\footnote{Easy MDP: \tiny{\url{https://github.com/zafarali/emdp}}}.
In the easiest setting, the agent starts at the bottom left and must travel to the top right in a world without obstacles.  We also include a double-slit maze, where a wall with two openings separates the starting position of the agent and the goal, and the standard four rooms domain. We use these environments to demonstrate diversity and mutual information in the $p(\pi,z)$ learned by our model. See Figure~\ref{fig:RL_manywalls}.

We analyze the behaviour of different policies induced by interpolating between a pair of latent variables,~i.e. we plot the state visitation frequencies of the policies $\pi(a|s,z_i)$, where $z_i = \alpha z_0 + (1-\alpha) z_1$, and $z_0, z_1 \sim p(z)$. We visualize the behaviour of policies trained in the double-slit gridworld in Figure.~\ref{fig:RL_maxent_interpolation}. When $\alpha=0$, policies tend to go through one of the slits. As $\alpha$ increases, policies gradually move toward the other slit. Around $\alpha=0.5$ trajectories may go through either slit.

\section{Discussion and Future Work}
We proposed a generative model for functions, and showed that it can learn to generate diverse sets of functions. The objectives optimized by our model make it suitable for Bayesian deep learning in a variety of contexts.

We demonstrated the behaviour of our model in toy supervised learning and reinforcement learning tasks. Our model can estimate predictive output uncertainty in simple regression tasks, as in Bayesian neural networks \cite{neal2012bayesian}. Our model can be used in reinforcement learning tasks, e.g.~we can maximize the entropy of a distribution over policies to obtain diverse policies that solve the task in distinct ways.

Future work will explore various applications of Bayesian learning, such as active learning and anomaly detection. In particular, our model may be useful in Thompson sampling schemes \cite{riquelme2018deep}. We will also explore further reinforcement learning tasks where learning a maximum entropy distribution over policies is useful. We anticipate that such a mechanism would be favourable to learn effective ways to solve exploration tasks in complex mazes \cite{beattie2016deepmind,gym_minigrid} or for quicker adaptation in continuous control domains \cite{lillicrap2015continuous,frans2017meta}. Explicit representations of a posterior over reward functions can also be useful in inverse reward design \cite{hadfield2017inverse}.

More abstract concerns also merit further consideration, e.g.~how to choose small but informative sets of points at which to evaluate partial functions $\hat{f}$ which most concisely capture the relationship between $f$ and $z$ in $p(f,z)$.

\bibliographystyle{icml2018}
\bibliography{main}

\begin{thebibliography}{40}
\providecommand{\natexlab}[1]{#1}
\providecommand{\url}[1]{\texttt{#1}}
\expandafter\ifx\csname urlstyle\endcsname\relax
  \providecommand{\doi}[1]{doi: #1}\else
  \providecommand{\doi}{doi: \begingroup \urlstyle{rm}\Url}\fi

\bibitem[Bachman et~al.(2014)Bachman, Alsharif, and Precup]{Bachman2014}
Bachman, Philip, Alsharif, Ouais, and Precup, Doina.
\newblock Learning with pseudo-ensembles.
\newblock \emph{Advances in Neural Information Processing Systems (NIPS)},
  2014.

\bibitem[Barber \& Agakov(2003)Barber and Agakov]{barber2003algorithm}
Barber, David and Agakov, Felix.
\newblock The im algorithm: a variational approach to information maximization.
\newblock \emph{International Conference on Neural Information Processing
  Systems}, 2003.

\bibitem[Beattie et~al.(2016)Beattie, Leibo, Teplyashin, Ward, Wainwright,
  Kuttler, Lefrancq, Green, Valdes, Sadik, et~al.]{beattie2016deepmind}
Beattie, Charles, Leibo, Joel~Z, Teplyashin, Denis, Ward, Tom, Wainwright,
  Marcus, Kuttler, Heinrich, Lefrancq, Andrew, Green, Simon, Valdes, Victor,
  Sadik, Amir, et~al.
\newblock Deepmind lab.
\newblock \emph{arXiv:1612.03801}, 2016.

\bibitem[Belghazi et~al.(2018)Belghazi, Baratin, Rajeswar, Ozair, Bengio,
  Courville, and Hjelm]{Belghazi2018}
Belghazi, Mohamed~Ishmael, Baratin, Aristide, Rajeswar, Sai, Ozair, Sherjil,
  Bengio, Yoshua, Courville, Aaron, and Hjelm, R~Devon.
\newblock Mine: Mutual information neural estimation.
\newblock \emph{International Conference on Machine Learning (ICML)}, 2018.

\bibitem[Bishop(2006)]{Bishop06}
Bishop, Christopher~M.
\newblock \emph{Pattern Recognition and Machine Learning}.
\newblock Springer, 2006.

\bibitem[Blundell et~al.(2015)Blundell, Cornebise, Kavukcuoglu, and
  Wierstra]{blundell2015}
Blundell, Charles, Cornebise, Julien, Kavukcuoglu, Koray, and Wierstra, Daan.
\newblock Weight uncertainty in neural networks.
\newblock \emph{arXiv:1505.05424}, 2015.

\bibitem[Dinh et~al.(2017)Dinh, Pascanu, Bengio, and Bengio]{dinh2017sharp}
Dinh, Laurent, Pascanu, Razvan, Bengio, Samy, and Bengio, Yoshua.
\newblock Sharp minima can generalize for deep nets.
\newblock \emph{arXiv:1703.04933}, 2017.

\bibitem[Eysenbach et~al.(2018)Eysenbach, Gupta, Ibarz, and Levine]{DIAYN}
Eysenbach, Benjamin, Gupta, Abhishek, Ibarz, Julian, and Levine, Sergey.
\newblock Diversity is all you need: Learning skills without a reward function.
\newblock \emph{arXiv:1802.06070}, 2018.

\bibitem[Finn \& Levine(2018)Finn and Levine]{Finn2018}
Finn, Chelsea and Levine, Sergey.
\newblock Meta-learning and universality: Deep representations and gradient
  descent can approximate any learning algorithm.
\newblock \emph{International Conference on Learning Representations (ICLR)},
  2018.

\bibitem[Florensa et~al.(2017)Florensa, Duan, and
  Abbeel]{florensa2017stochastic}
Florensa, Carlos, Duan, Yan, and Abbeel, Pieter.
\newblock Stochastic neural networks for hierarchical reinforcement learning.
\newblock \emph{International Conference on Learning Representations (ICLR)},
  2017.

\bibitem[Fortunato et~al.(2017)Fortunato, Azar, Piot, Menick, Osband, Graves,
  Mnih, Munos, Hassabis, Pietquin, Blundell, and Legg]{fortunato2017}
Fortunato, Meire, Azar, Mohammad~Gheshlaghi, Piot, Bilal, Menick, Jacob,
  Osband, Ian, Graves, Alex, Mnih, Vlad, Munos, Remi, Hassabis, Demis,
  Pietquin, Olivier, Blundell, Charles, and Legg, Shane.
\newblock Noisy networks for exploration.
\newblock \emph{arXiv:1706.10295}, 2017.

\bibitem[Frans et~al.(2017)Frans, Ho, Chen, Abbeel, and
  Schulman]{frans2017meta}
Frans, Kevin, Ho, Jonathan, Chen, Xi, Abbeel, Pieter, and Schulman, John.
\newblock Meta learning shared hierarchies.
\newblock \emph{arXiv:1710.09767}, 2017.

\bibitem[Gal \& Ghahramani(2016)Gal and Ghahramani]{gal2016dropout}
Gal, Yarin and Ghahramani, Zoubin.
\newblock Dropout as a bayesian approximation: Representing model uncertainty
  in deep learning.
\newblock \emph{International Conference on Machine Learning (ICML)}, 2016.

\bibitem[Gal et~al.(2017)Gal, Islam, and Ghahramani]{gal2017deep}
Gal, Yarin, Islam, Riashat, and Ghahramani, Zoubin.
\newblock Deep bayesian active learning with image data.
\newblock \emph{International Conference on Machine Learning (ICML)}, 2017.

\bibitem[Garnelo et~al.(2018)Garnelo, Schwarz, Rosenbaum, Viola, Rezende,
  Eslami, and Teh]{garnelo2018}
Garnelo, Marta, Schwarz, Jonathan, Rosenbaum, Dan, Viola, Fabio, Rezende,
  Danilo~J., Eslami, S.M.~Ali, and Teh, Yee~Whye.
\newblock Neural processes.
\newblock \emph{arXiv:1807.01622}, 2018.

\bibitem[Gupta et~al.(2018{\natexlab{a}})Gupta, Mendonca, Liu, Abbeel, and
  Levine]{MAESN}
Gupta, Abhishek, Mendonca, Russell, Liu, Yuxuan, Abbeel, Pieter, and Levine,
  Sergey.
\newblock Meta-reinforcement learning of structured exploration strategies.
\newblock \emph{arxiv:1802.07245}, 2018{\natexlab{a}}.

\bibitem[Gupta et~al.(2018{\natexlab{b}})Gupta, Mendonca, Liu, Abbeel, and
  Levine]{gupta2018meta}
Gupta, Abhishek, Mendonca, Russell, Liu, YuXuan, Abbeel, Pieter, and Levine,
  Sergey.
\newblock Meta-reinforcement learning of structured exploration strategies.
\newblock \emph{arXiv:1802.07245}, 2018{\natexlab{b}}.

\bibitem[Haarnoja et~al.(2018{\natexlab{a}})Haarnoja, Hartikainen, Abbeel, and
  Levine]{latentpolicies}
Haarnoja, Tuomas, Hartikainen, Kristian, Abbeel, Pieter, and Levine, Sergey.
\newblock Latent space policies for hierarchical reinforcement learning.
\newblock \emph{arXiv:1804.02808}, 2018{\natexlab{a}}.

\bibitem[Haarnoja et~al.(2018{\natexlab{b}})Haarnoja, Zhou, Abbeel, and
  Levine]{SAC}
Haarnoja, Tuomas, Zhou, Aurick, Abbeel, Pieter, and Levine, Sergey.
\newblock Soft actor-critic: Off-policy maximum entropy deep reinforcement
  learning with a stochastic actor.
\newblock \emph{arXiv:1801.01290}, 2018{\natexlab{b}}.

\bibitem[Hadfield-Menell et~al.(2017)Hadfield-Menell, Milli, Abbeel, Russell,
  and Dragan]{hadfield2017inverse}
Hadfield-Menell, Dylan, Milli, Smitha, Abbeel, Pieter, Russell, Stuart~J, and
  Dragan, Anca.
\newblock Inverse reward design.
\newblock \emph{Advances in Neural Information Processing Systems (NIPS)},
  2017.

\bibitem[Hausman et~al.(2018)Hausman, Springenberg, Wang, Heess, and
  Riedmiller]{hausman2018learning}
Hausman, Karol, Springenberg, Jost~Tobias, Wang, Ziyu, Heess, Nicolas, and
  Riedmiller, Martin.
\newblock Learning an embedding space for transferable robot skills.
\newblock \emph{International Conference on Learning Representations}, 2018.

\bibitem[Kendall \& Gal(2017)Kendall and Gal]{kendall2017uncertainties}
Kendall, Alex and Gal, Yarin.
\newblock What uncertainties do we need in bayesian deep learning for computer
  vision?
\newblock \emph{Advances in Neural Information Processing Systems (NIPS)},
  2017.

\bibitem[Kingma \& Welling(2013)Kingma and Welling]{kingma2013auto}
Kingma, Diederik~P and Welling, Max.
\newblock Auto-encoding variational bayes.
\newblock \emph{arXiv:1312.6114}, 2013.

\bibitem[Krueger et~al.(2017)Krueger, Huang, Islam, Turner, Lacoste, and
  Courville]{BHNs}
Krueger, David, Huang, Chin{-}Wei, Islam, Riashat, Turner, Ryan, Lacoste,
  Alexandre, and Courville, Aaron~C.
\newblock Bayesian hypernetworks.
\newblock \emph{arXiv:1710.04759}, 2017.

\bibitem[Lillicrap et~al.(2015)Lillicrap, Hunt, Pritzel, Heess, Erez, Tassa,
  Silver, and Wierstra]{lillicrap2015continuous}
Lillicrap, Timothy~P, Hunt, Jonathan~J, Pritzel, Alexander, Heess, Nicolas,
  Erez, Tom, Tassa, Yuval, Silver, David, and Wierstra, Daan.
\newblock Continuous control with deep reinforcement learning.
\newblock \emph{arXiv:1509.02971}, 2015.

\bibitem[Louizos \& Welling(2017)Louizos and
  Welling]{louizos2017multiplicative}
Louizos, Christos and Welling, Max.
\newblock Multiplicative normalizing flows for variational bayesian neural
  networks.
\newblock \emph{arXiv:1703.01961}, 2017.

\bibitem[Maxime Chevalier-Boisvert(2018)]{gym_minigrid}
Maxime Chevalier-Boisvert, Lucas~Willems.
\newblock Minimalistic gridworld environment for openai gym.
\newblock \emph{GitHub repository}, 2018.

\bibitem[Nachum et~al.(2017)Nachum, Norouzi, Xu, and
  Schuurmans]{nachum2017bridging}
Nachum, Ofir, Norouzi, Mohammad, Xu, Kelvin, and Schuurmans, Dale.
\newblock Bridging the gap between value and policy based reinforcement
  learning.
\newblock \emph{Advances in Neural Information Processing Systems (NIPS)},
  2017.

\bibitem[Neal(2012)]{neal2012bayesian}
Neal, Radford~M.
\newblock \emph{Bayesian learning for neural networks}, volume 118.
\newblock Springer Science and Business Media, 2012.

\bibitem[Neu et~al.(2017)Neu, Jonsson, and Gomez]{neu2017unified}
Neu, Gergely, Jonsson, Anders, and Gomez, Vicenc.
\newblock A unified view of entropy-regularized markov decision processes.
\newblock \emph{arXiv:1705.07798}, 2017.

\bibitem[Osband et~al.(2016)Osband, Blundell, Pritzel, and
  Van~Roy]{osband2016deep}
Osband, Ian, Blundell, Charles, Pritzel, Alexander, and Van~Roy, Benjamin.
\newblock Deep exploration via bootstrapped dqn.
\newblock \emph{Advances in neural information processing systems (NIPS)},
  2016.

\bibitem[Pawlowski et~al.(2017)Pawlowski, Rajchl, and
  Glocker]{pawlowski2017implicit}
Pawlowski, Nick, Rajchl, Martin, and Glocker, Ben.
\newblock Implicit weight uncertainty in neural networks.
\newblock \emph{arXiv:1711.01297}, 2017.

\bibitem[Plappert et~al.(2017)Plappert, Houthooft, Dhariwal, Sidor, Chen, Chen,
  Asfour, Abbeel, and Andrychowicz]{plappert2017parameter}
Plappert, Matthias, Houthooft, Rein, Dhariwal, Prafulla, Sidor, Szymon, Chen,
  Richard~Y, Chen, Xi, Asfour, Tamim, Abbeel, Pieter, and Andrychowicz, Marcin.
\newblock Parameter space noise for exploration.
\newblock \emph{arXiv:1706.01905}, 2017.

\bibitem[Rasmussen \& Williams(2006)Rasmussen and Williams]{gaussianprocesses}
Rasmussen, Carl~Edward and Williams, Christopher K.~I.
\newblock Gaussian processes for machine learning.
\newblock \emph{Adaptive computation and machine learning}, 2006.

\bibitem[Riquelme et~al.(2018)Riquelme, Tucker, and Snoek]{riquelme2018deep}
Riquelme, Carlos, Tucker, George, and Snoek, Jasper.
\newblock Deep bayesian bandits showdown: An empirical comparison of bayesian
  deep networks for thompson sampling.
\newblock \emph{arXiv:1802.09127}, 2018.

\bibitem[Russo et~al.(2017)Russo, Roy, Kazerouni, and Osband]{Russo17}
Russo, Daniel, Roy, Benjamin~Van, Kazerouni, Abbas, and Osband, Ian.
\newblock A tutorial on thompson sampling.
\newblock \emph{arXiv:1707.02038}, 2017.

\bibitem[Thompson(1933)]{thompson1933likelihood}
Thompson, William~R.
\newblock On the likelihood that one unknown probability exceeds another in
  view of the evidence of two samples.
\newblock \emph{Biometrika}, 1933.

\bibitem[Williams(1992)]{williams1992simple}
Williams, Ronald~J.
\newblock Simple statistical gradient-following algorithms for connectionist
  reinforcement learning.
\newblock \emph{Reinforcement Learning}, 1992.

\bibitem[Xie et~al.(2007)Xie, Lord, and Zhang]{xie2007predicting}
Xie, Yuanchang, Lord, Dominique, and Zhang, Yunlong.
\newblock Predicting motor vehicle collisions using bayesian neural network
  models: An empirical analysis.
\newblock \emph{Accident Analysis and Prevention}, 2007.

\bibitem[Zaheer et~al.(2017)Zaheer, Kottur, Ravanbakhsh, Poczos, Salakhutdinov,
  and Smola]{Zaheer2017}
Zaheer, Manzil, Kottur, Satwik, Ravanbakhsh, Siamak, Poczos, Barnabas,
  Salakhutdinov, Ruslan, and Smola, Alexander.
\newblock Deep sets.
\newblock \emph{Advances in Neural Information Processing Systems (NIPS)},
  2017.

\end{thebibliography}

\clearpage
\appendix
\section{Appendix}
\subsection{More Model Details}

\begin{figure}[H]
	\centering
    \includegraphics[scale=0.35]{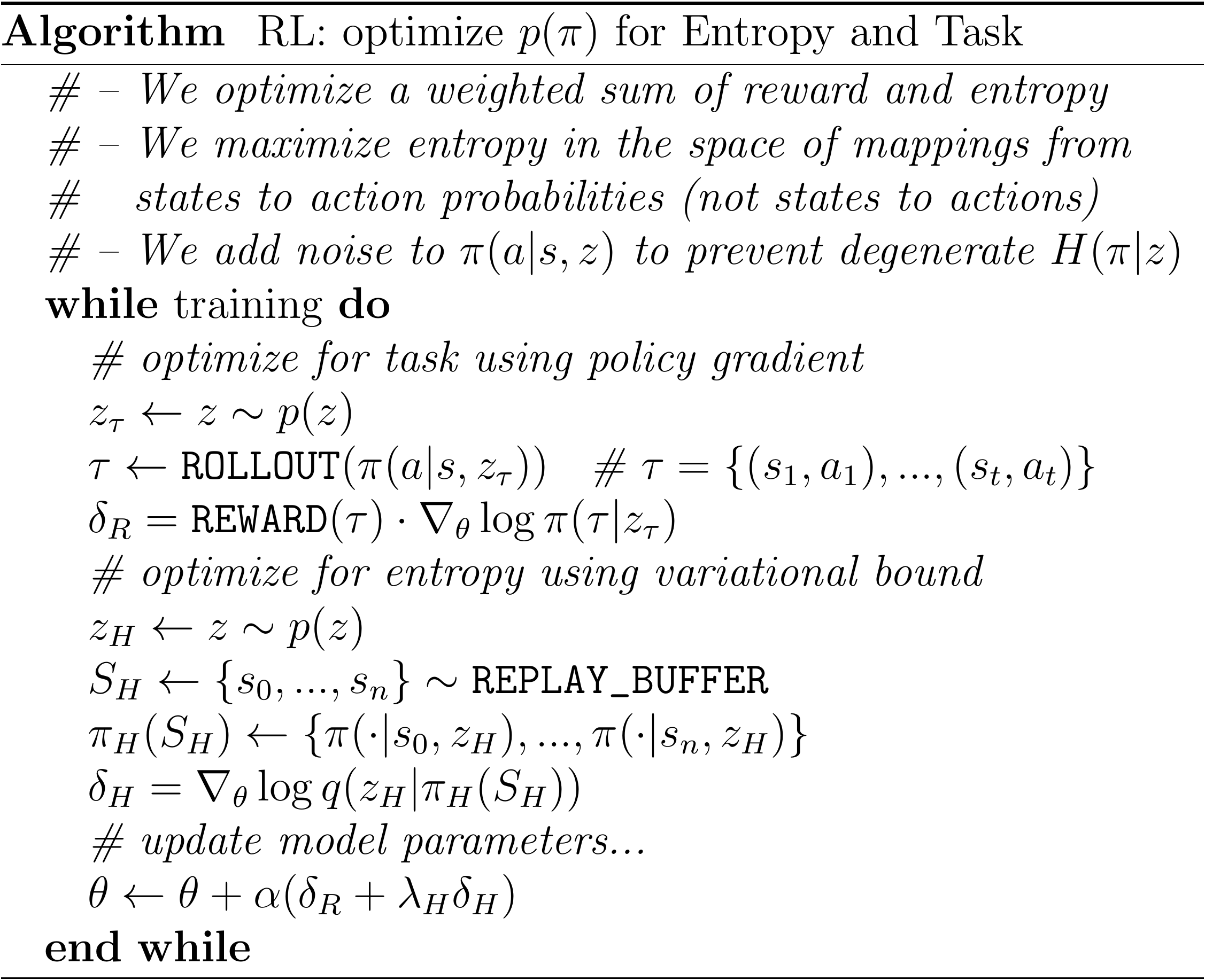}
    \caption{Pseudo-code for the main RL training loop when maximizing a sum of entropy and expected rewards using our model.}
\end{figure}

\begin{figure}[H]
	\centering
    \includegraphics[scale=0.35]{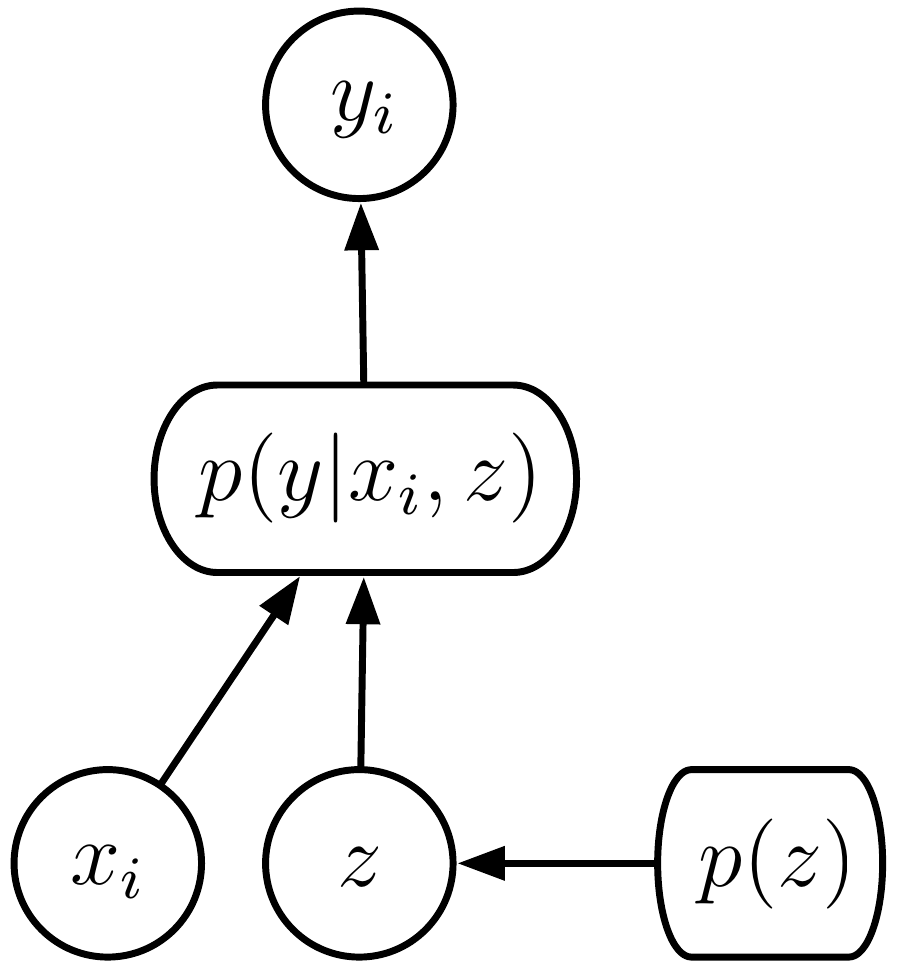} \\
    \vspace{0.3cm}
    \includegraphics[scale=0.35]{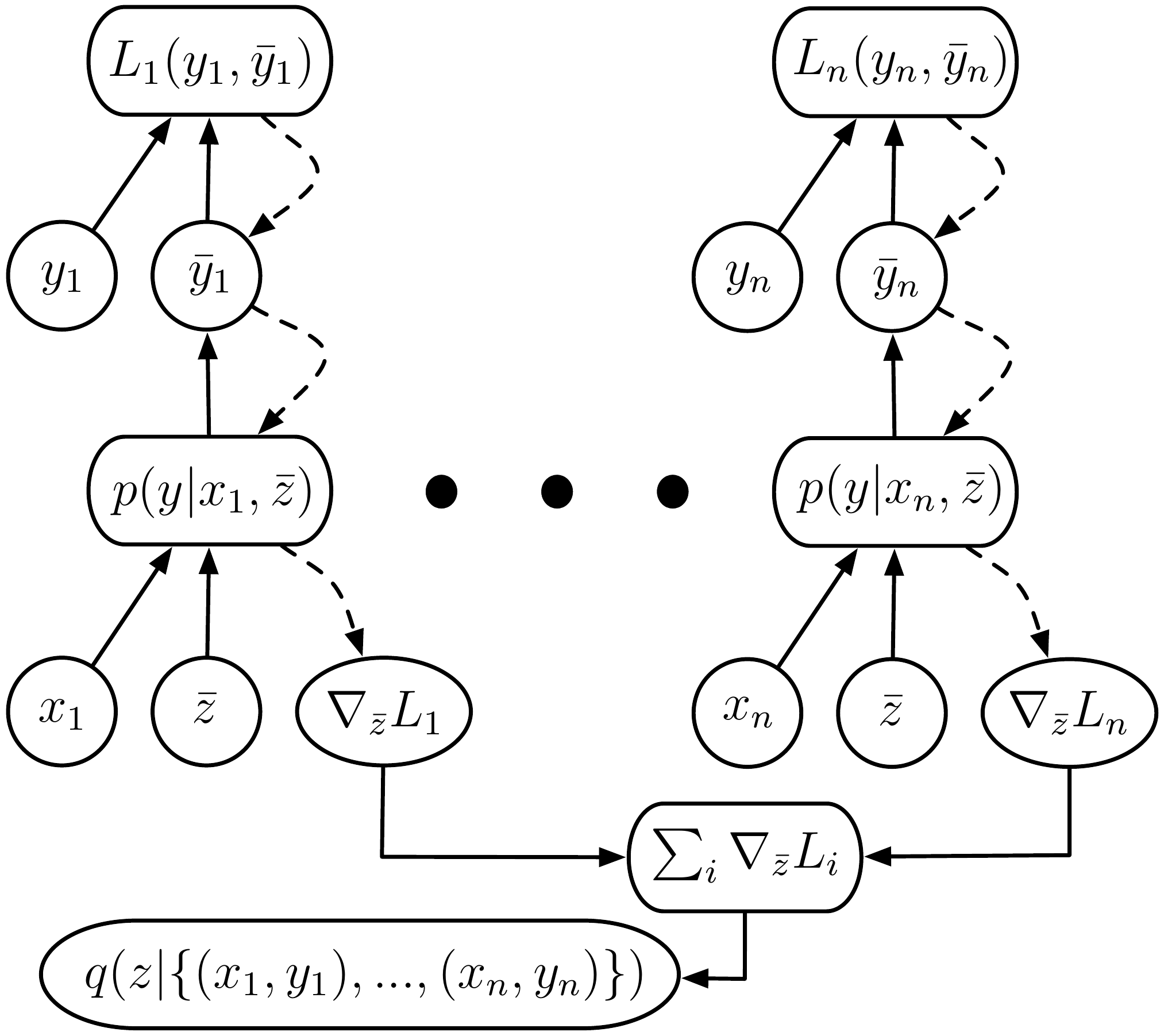}
    \caption{\textbf{Top}: the basic decoder architecture for our model. \textbf{Bottom}: the basic encoder architecture for our model. Circles denote variables, rounded boxes denote computations, solid lines denote typical forward-prop information flow, and dotted lines indicate backprop information flow that is part of the primary computation.}
\end{figure}

\subsection{More Reinforcement Learning Experiments}

\begin{figure}[H]
    \centering
    \footnotesize{(a) $\lambda_{\ent}=0$~~~~~~~~~~~~~~~~~~~~~~~~~(b) $\lambda_{\ent}=1$}\\
    \includegraphics[scale=0.16]{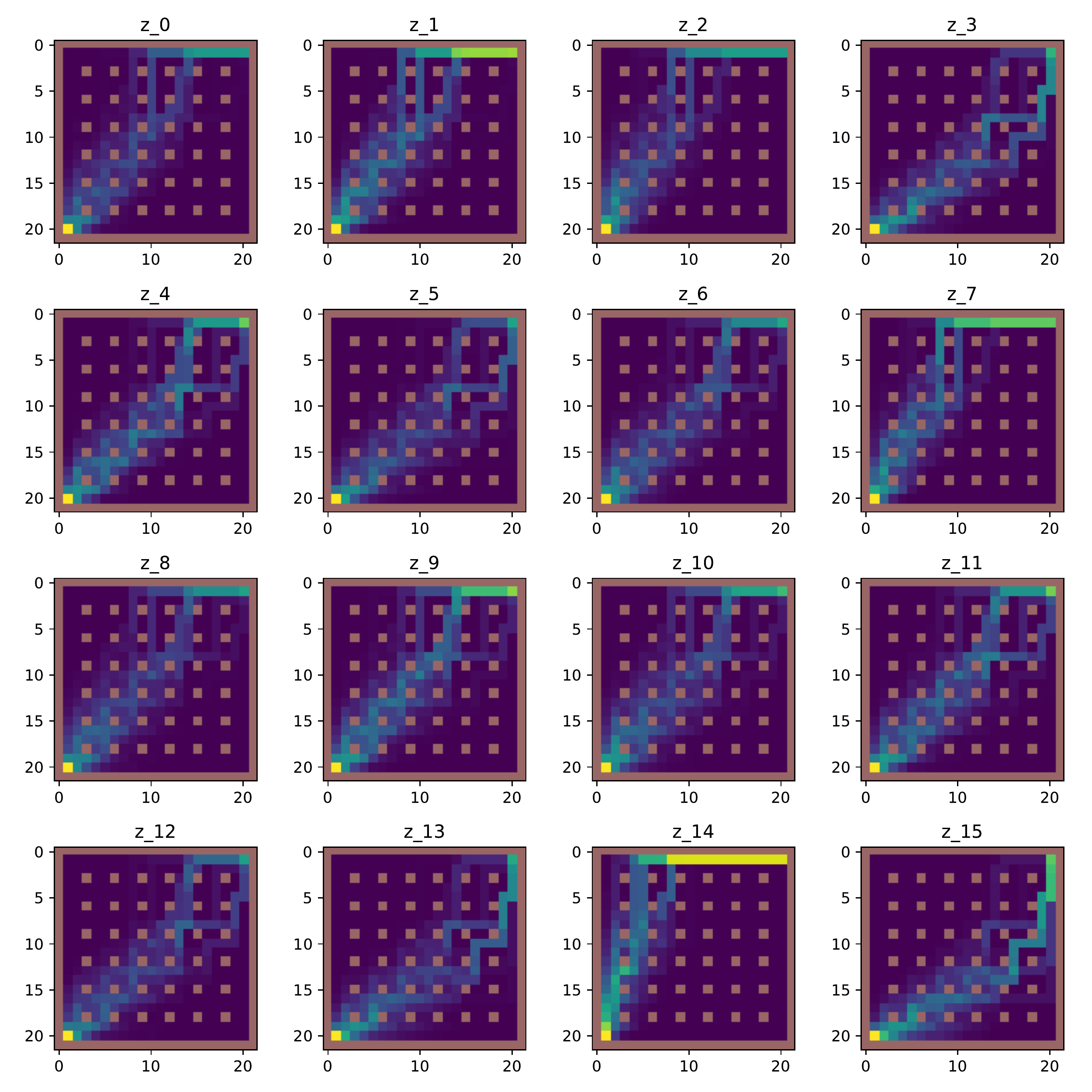}
    \includegraphics[scale=0.16]{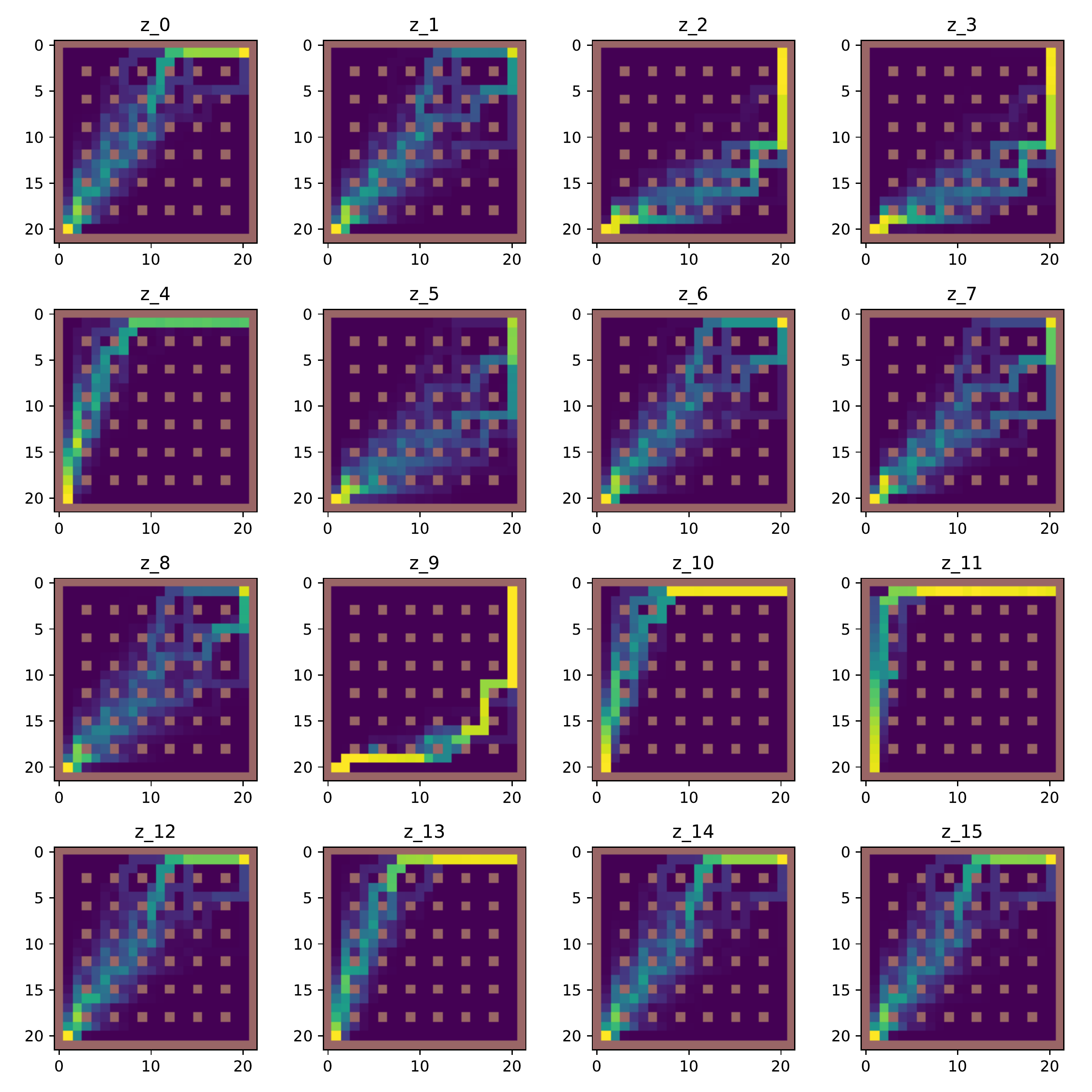}
    \caption{\textbf{Sampled policies in the Pachinko world} for an agent trained (a) without and (b) with entropy maximization. Both agents had latent-conditioned policies, but (a) did not explicitly maximize $I(\pi;z)$. The agent trained with entropy maximization learned a more diverse distribution of policies with larger $I(\pi; z)$.}
    \label{fig:RL_PachinkoWorlds_lament1}
\end{figure}

\begin{figure}[H]
    \begin{center}
    \includegraphics[scale=0.16]{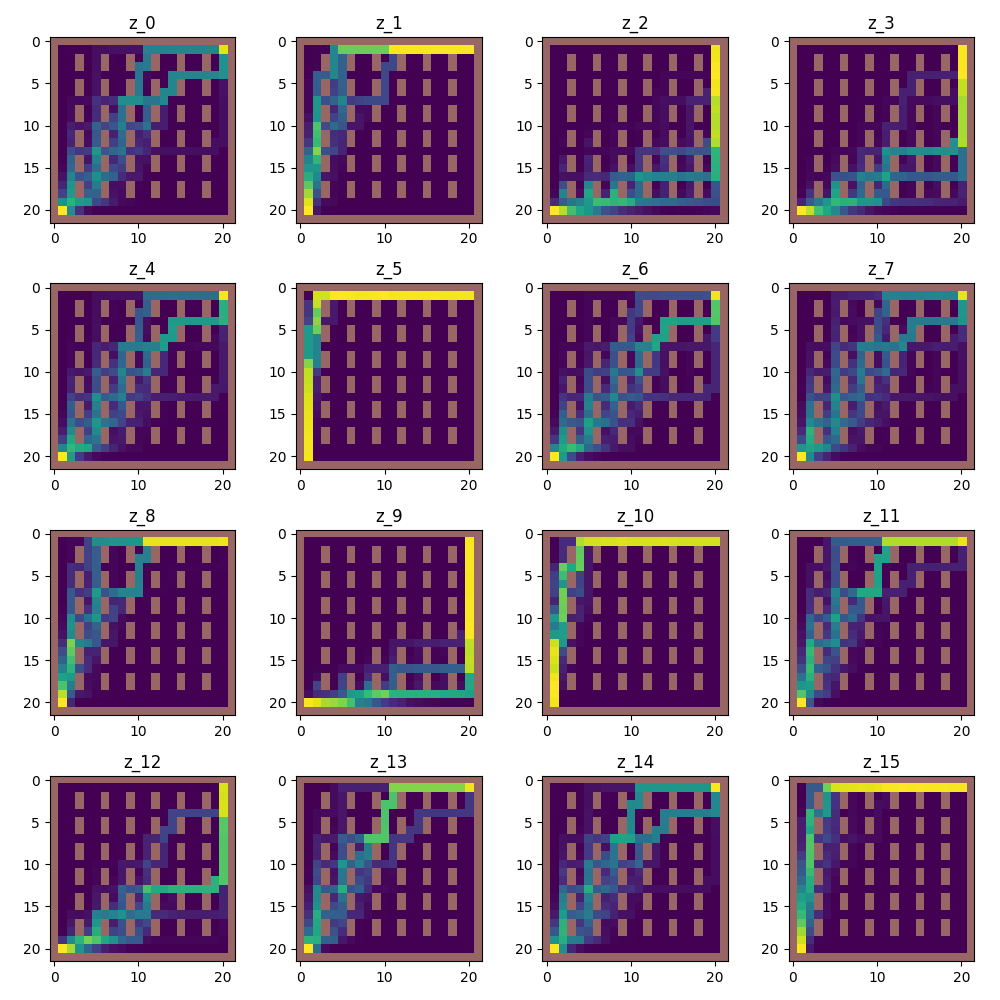}
    \includegraphics[scale=0.16]{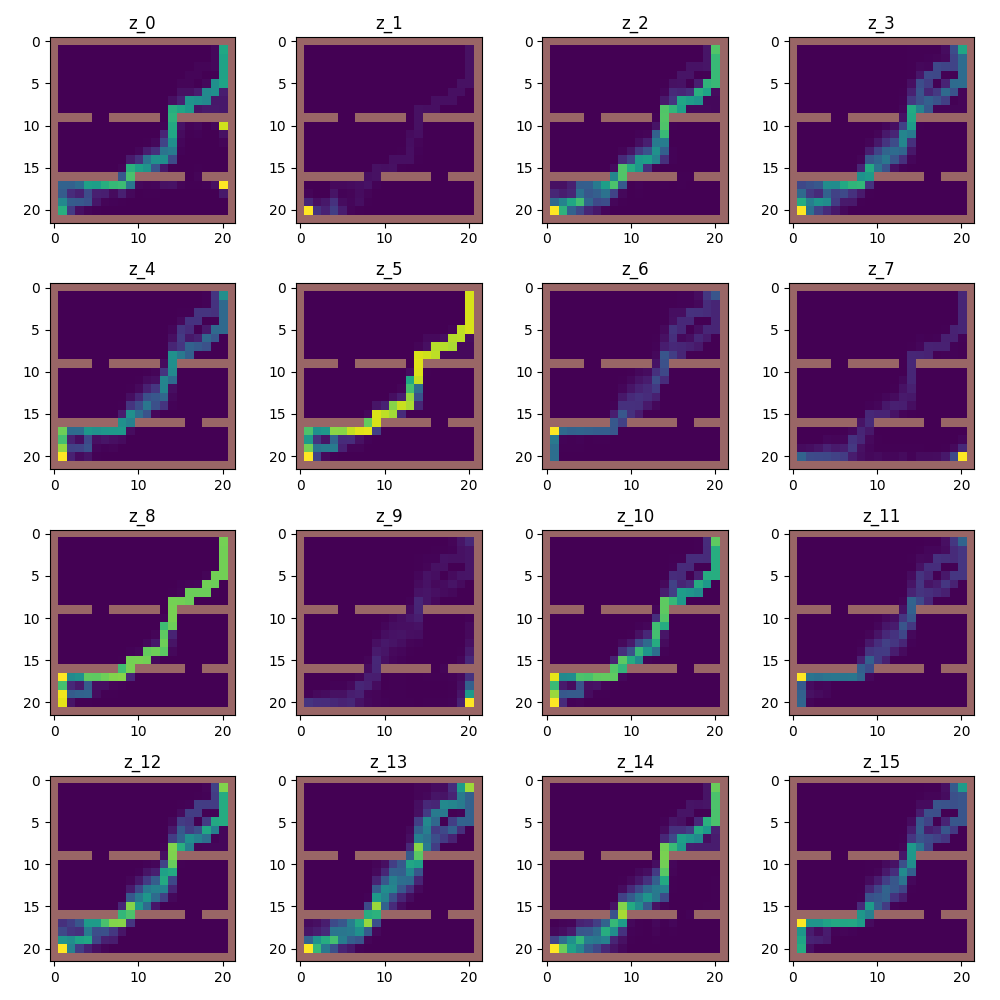}
    \includegraphics[scale=0.16]{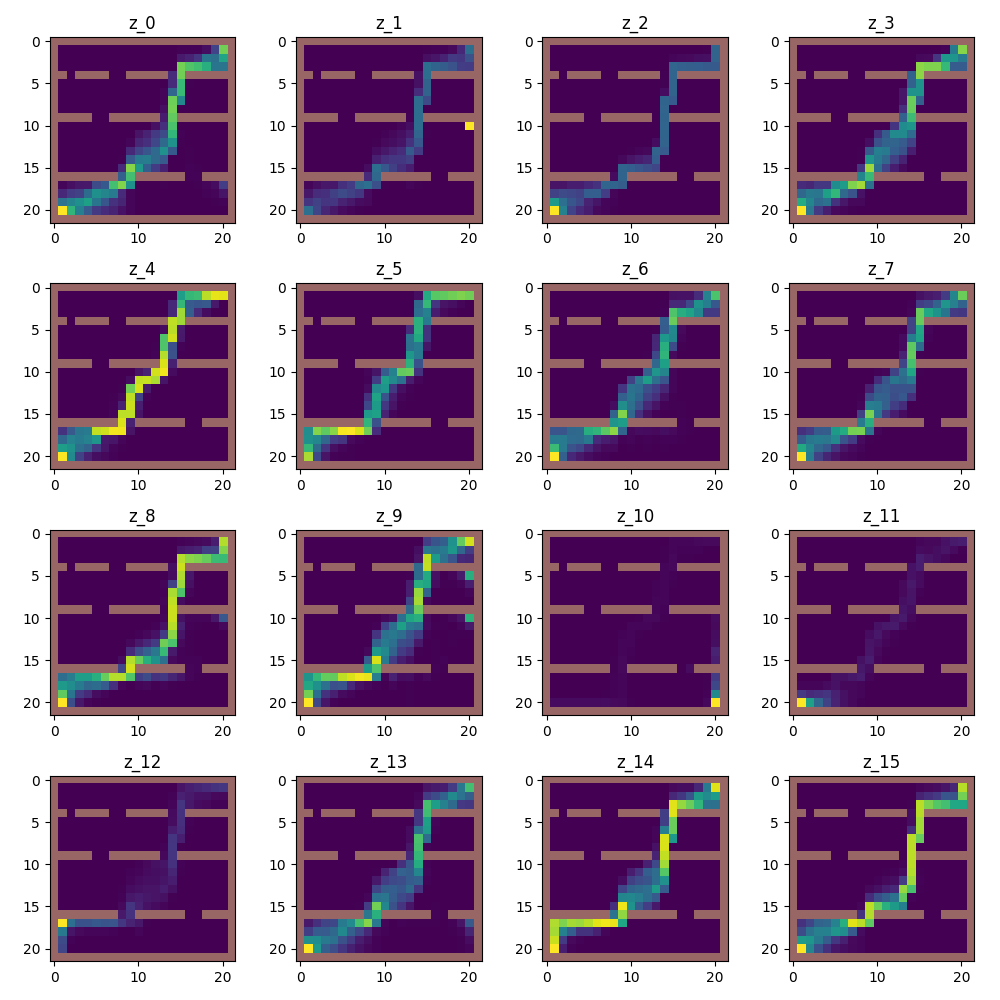}
    \includegraphics[scale=0.16]{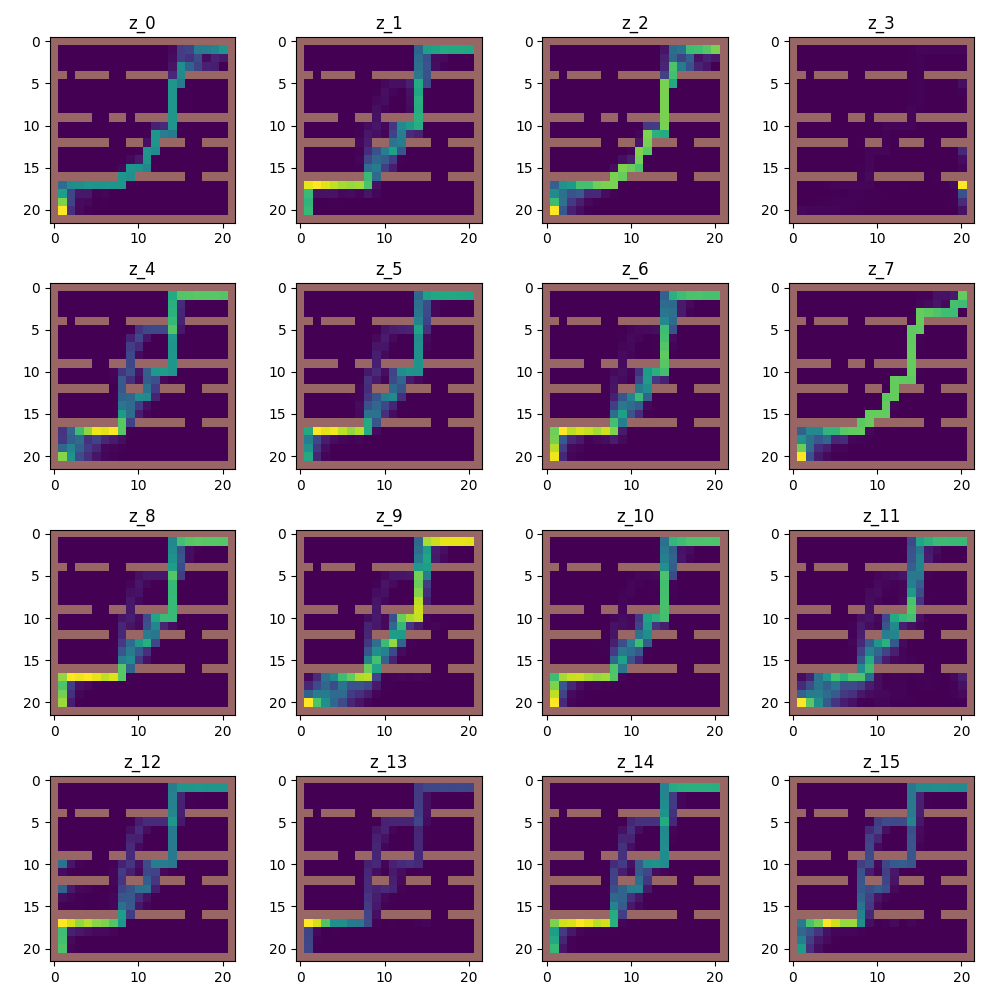}
    \caption{Policy distributions in several more domains: (a) Complex Pachinko (b) Double double slits (c) Many slits domain (d) Even more slits.}
    \label{fig:other_RL_tasks}
    \end{center}
\end{figure}

\subsection{More Regression Experiments}

\subsubsection{CO2 Dataset}
\begin{figure}[H]
    \centering
    \includegraphics[scale=0.25]{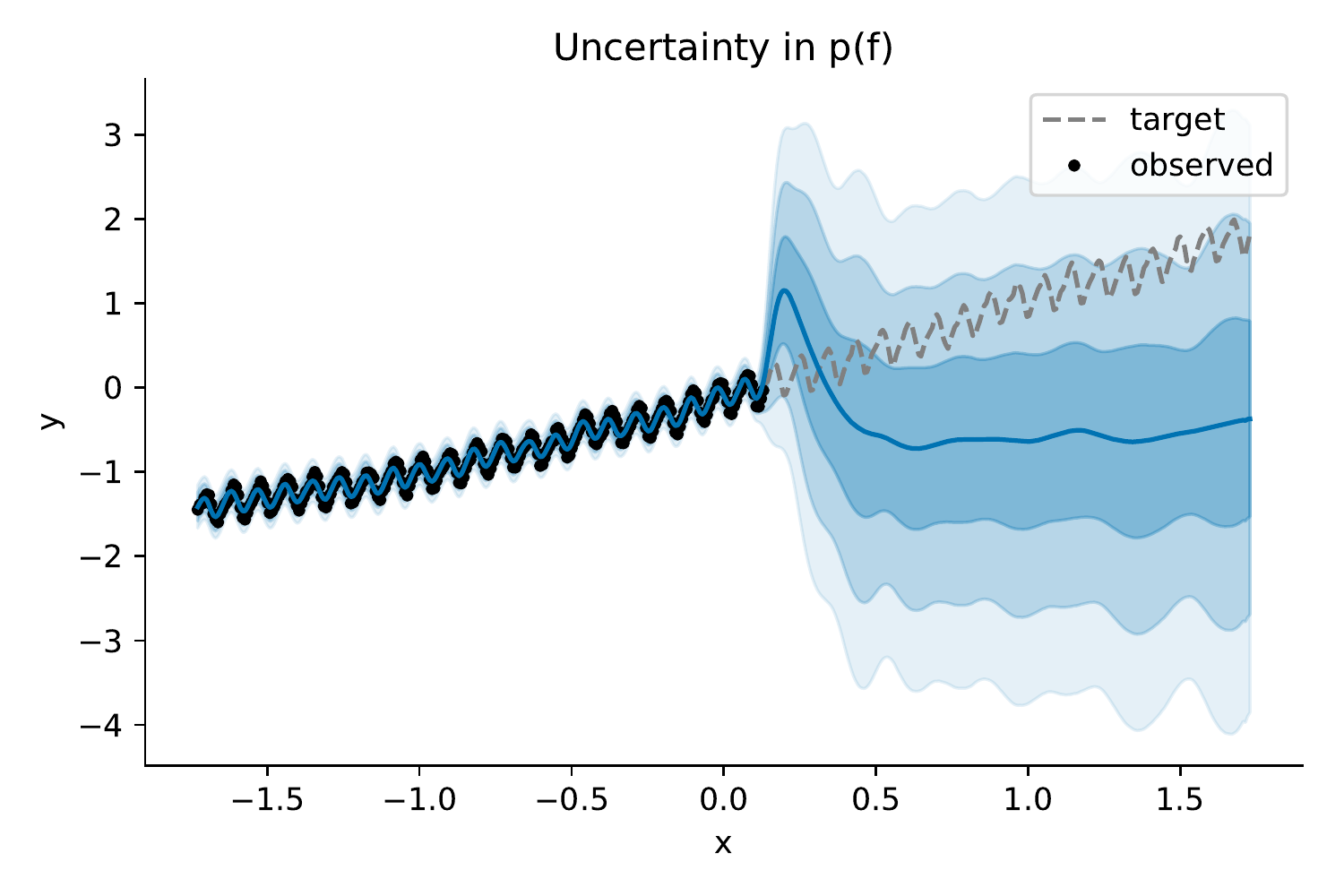} 
    \includegraphics[scale=0.2]{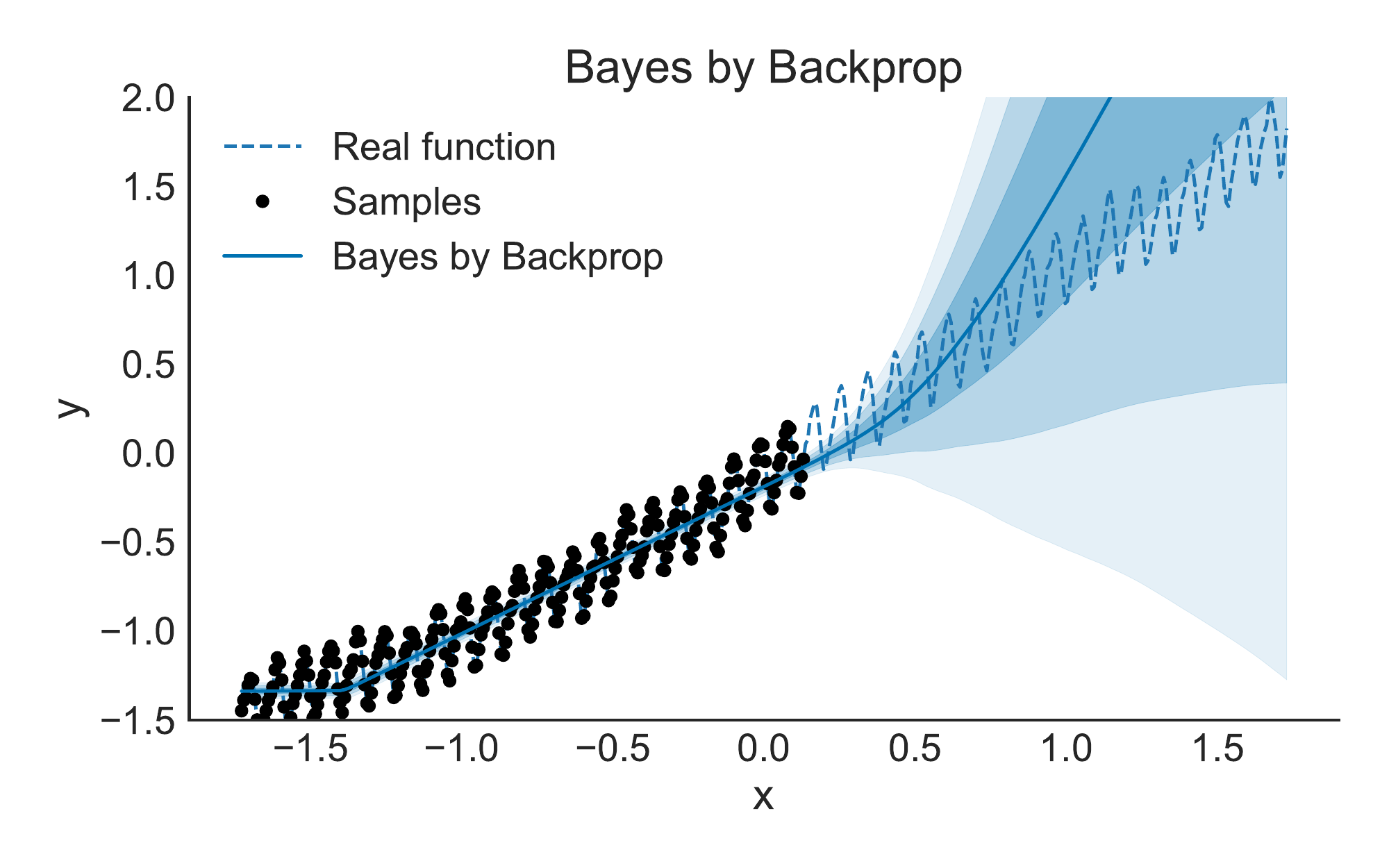}
    \includegraphics[scale=0.2]{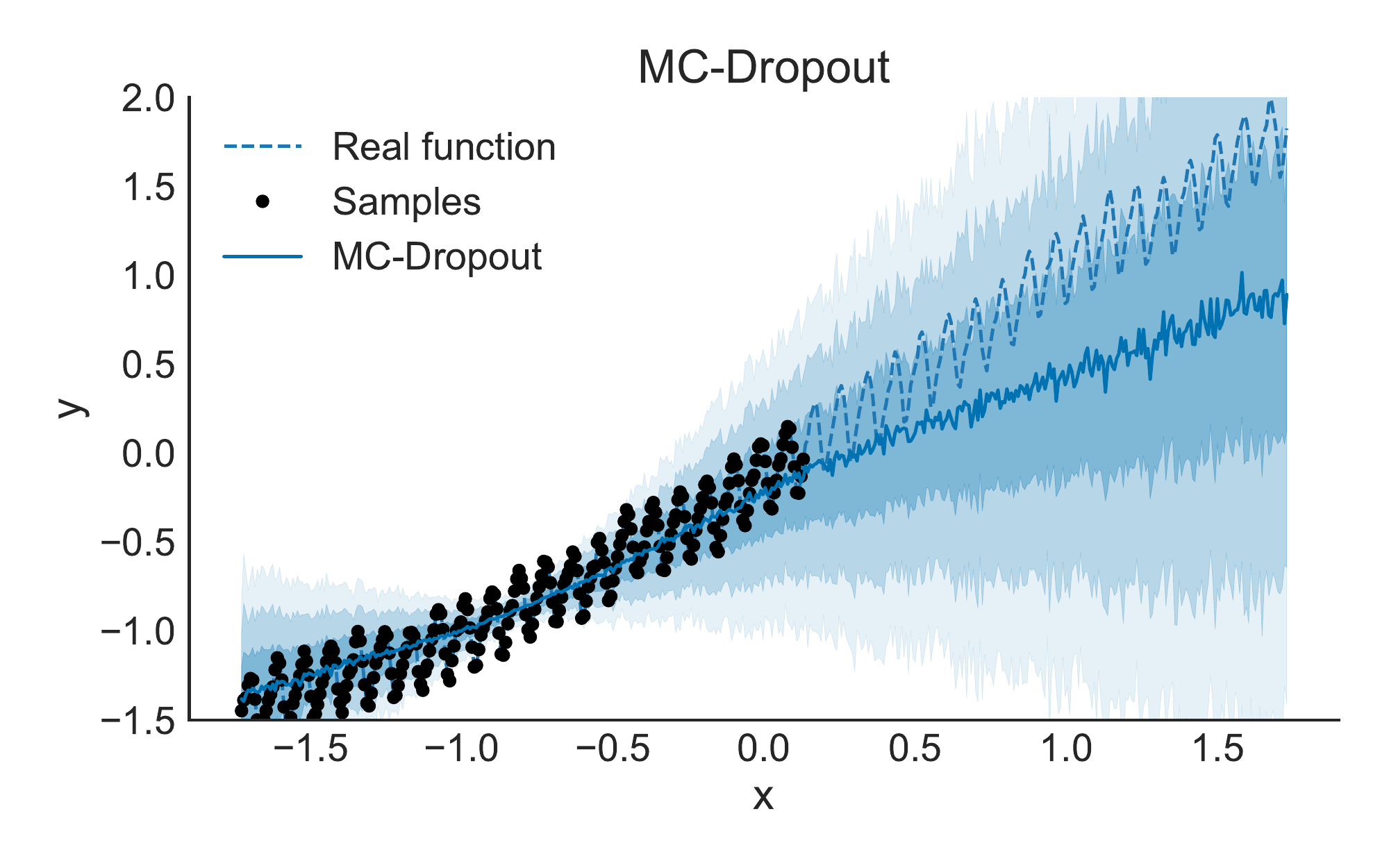}
    \includegraphics[scale=0.2]{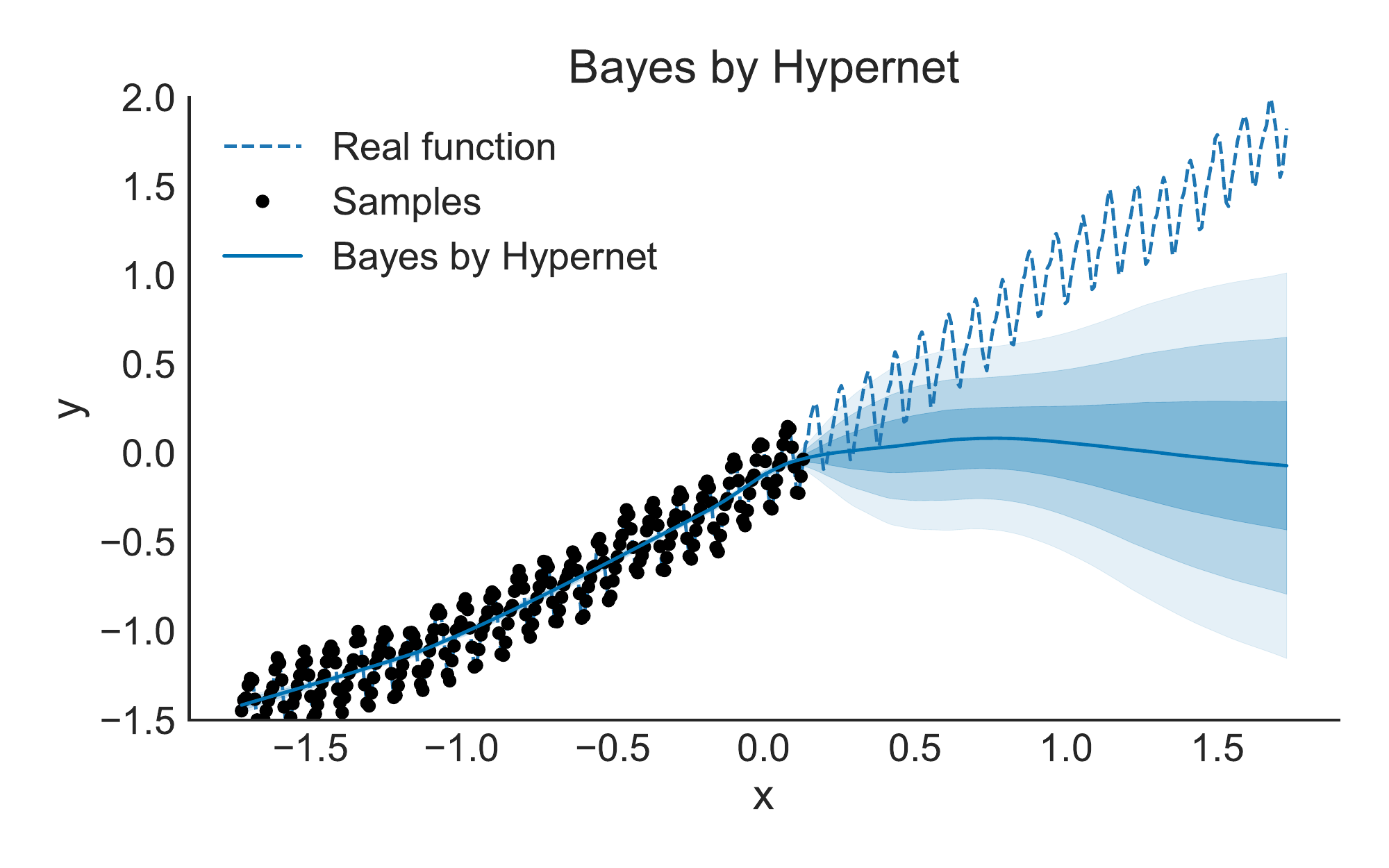}    
    \caption{Compare our model and baselines on CO2 data.}
    \label{fig:co2_comparison1}
\end{figure}

\subsubsection{Centered CO2 Dataset}
\begin{figure}[H]
    \centering
    \includegraphics[scale=0.25]{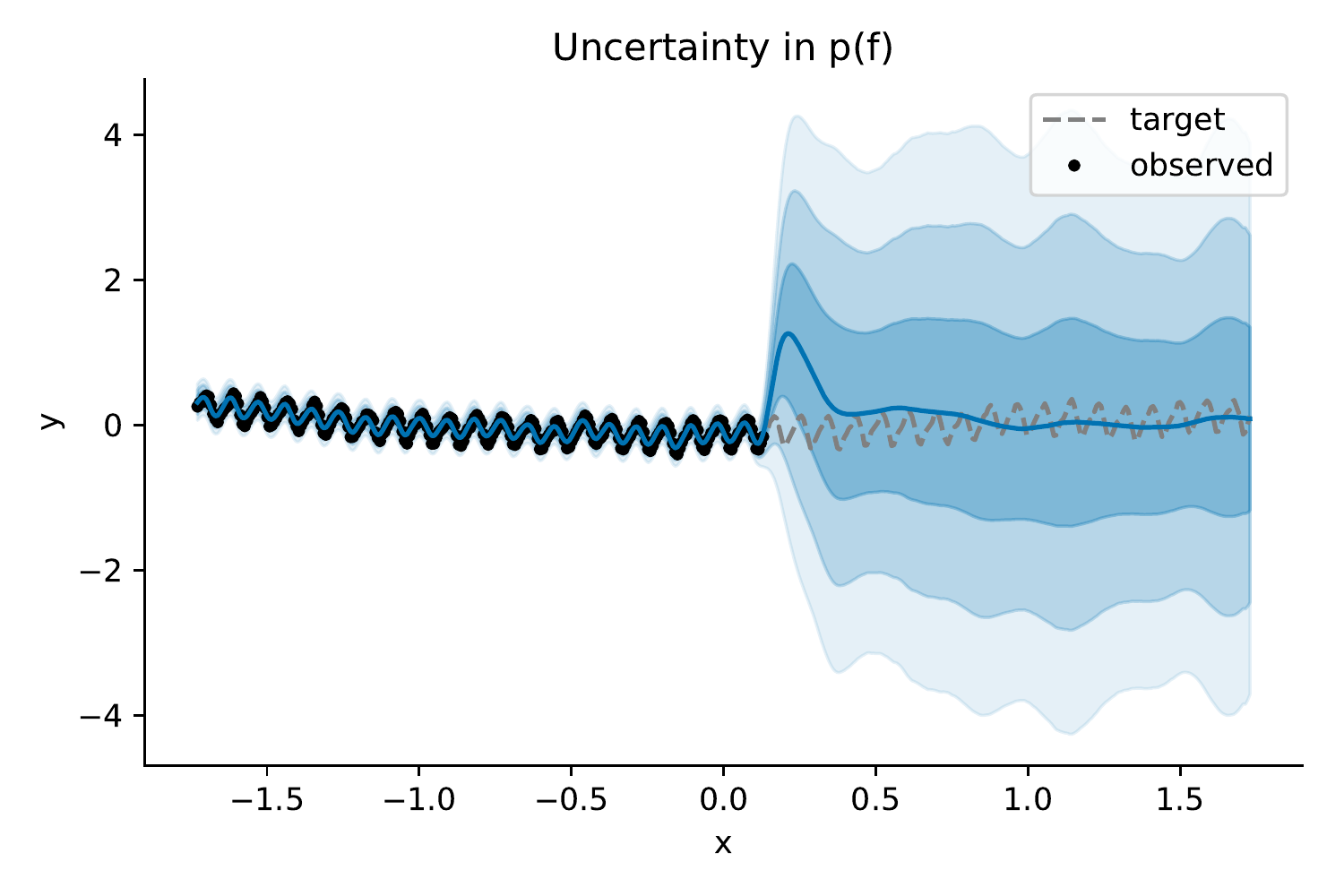}
    \includegraphics[scale=0.2]{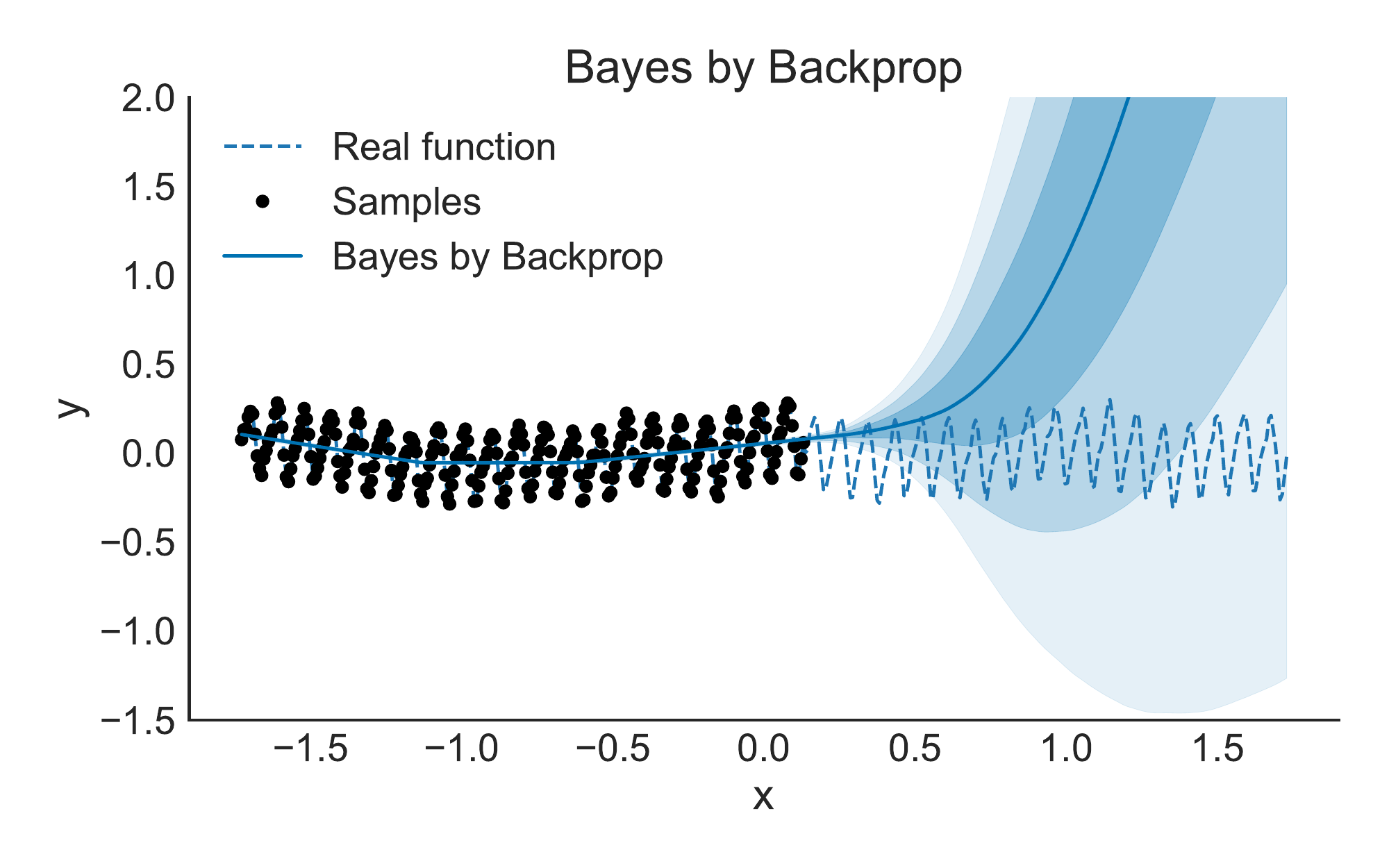}
    \includegraphics[scale=0.2]{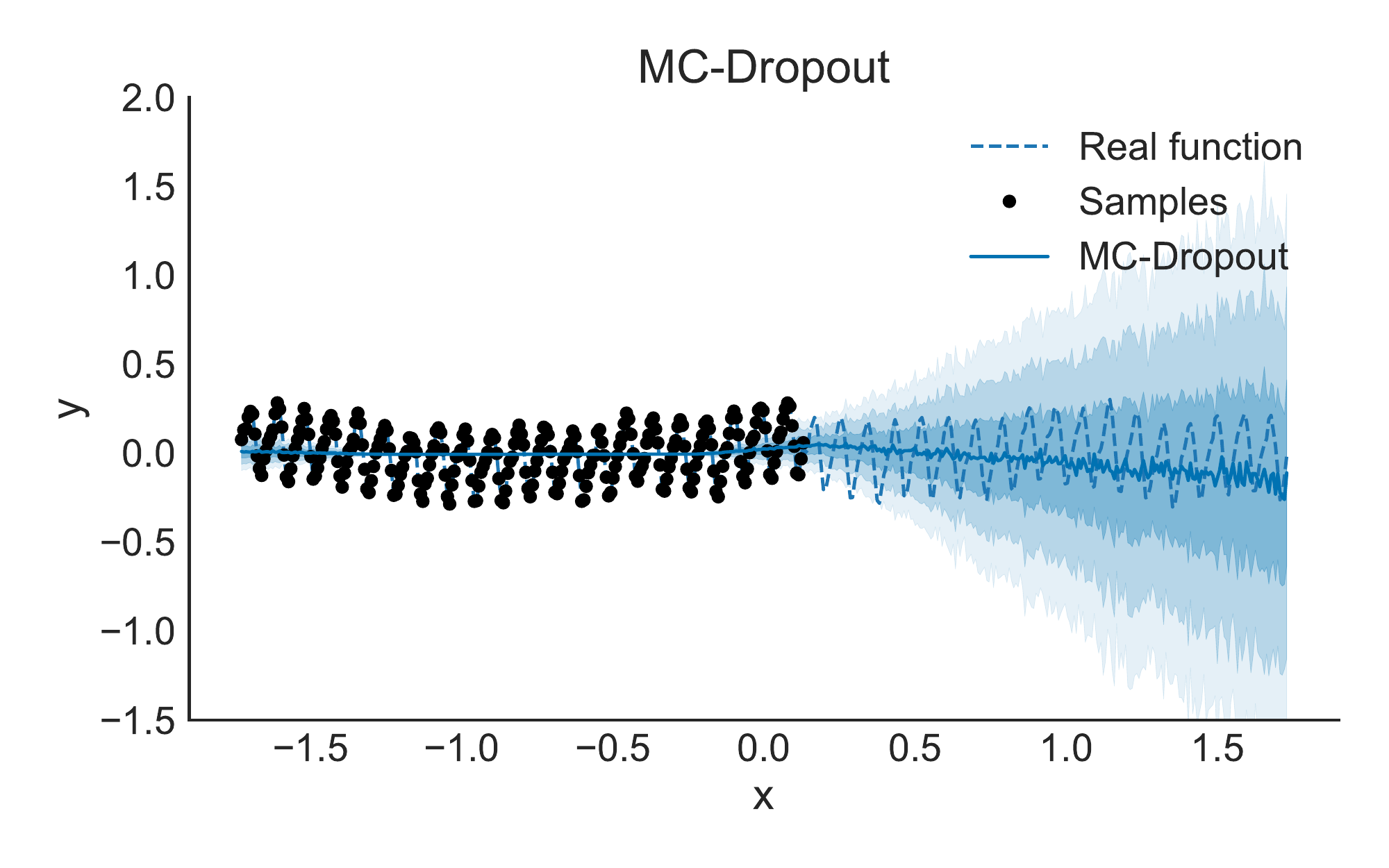}
    \includegraphics[scale=0.2]{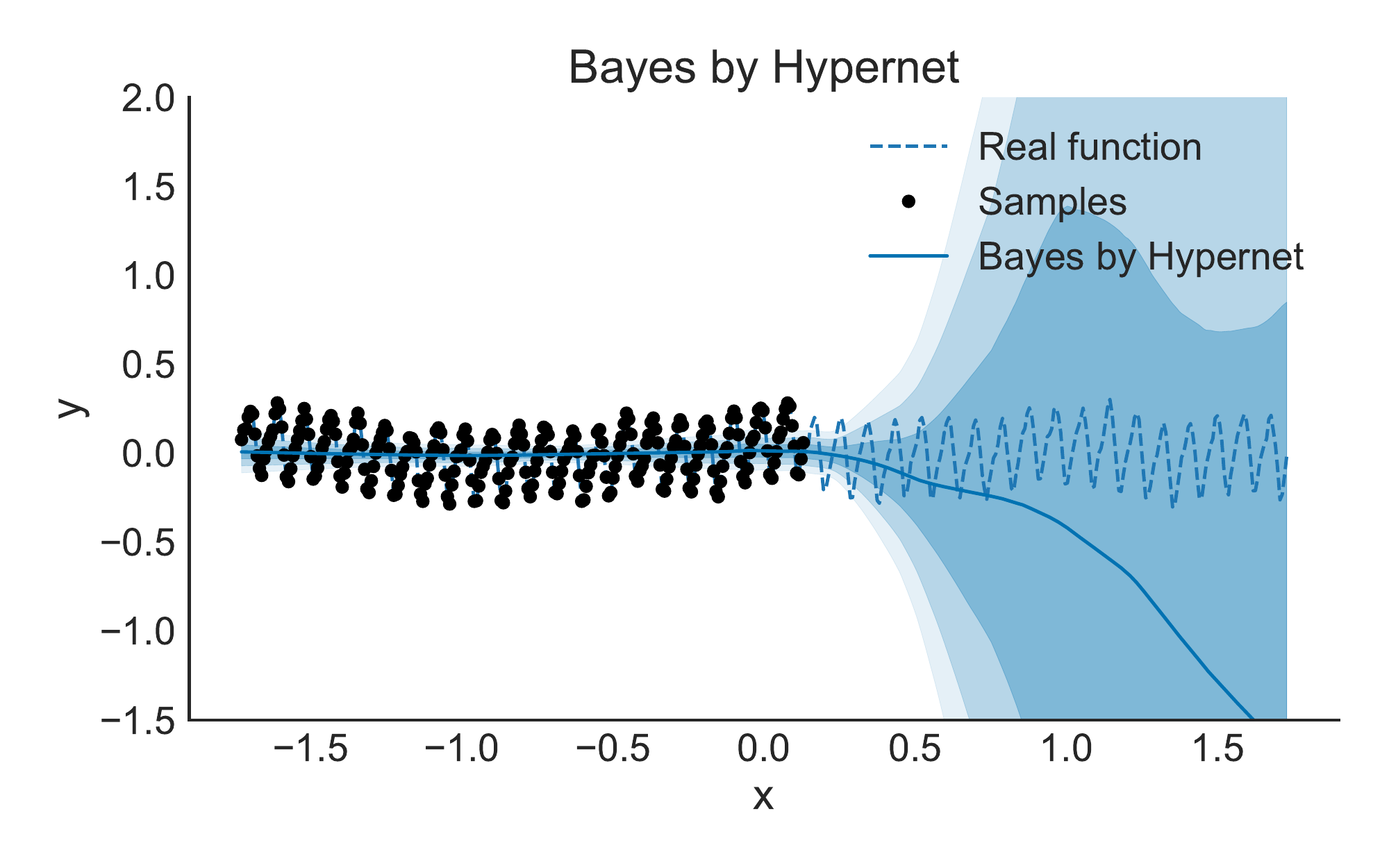}    
    \caption{Compare our model and baselines on centered CO2 data.}
    \label{fig:co2_comparison2}
\end{figure}


\clearpage
\section{Dynamic Discretization Bound}
\label{sec:dd_bound_appendix}
For now, we only discuss the case of discrete $z$s. This covers anything running in finite time on a digital computer, which includes most practical cases.

\subsection{Detailed Description}
We can construct an alternative variational lower bound on $H(f)$ by considering:
\begin{equation}
H(f) = I(f; z) + H(f|z).
\end{equation}
We can place a lower bound on $I_k(f;z)$ for any $kz$-subset of the full joint $p(f,z)$,
where ``$kz$-subset'' denotes a distribution $p_k(f,z)$ which only includes $(f,z)$ samples from $p(f,z)$ that touch a particular set of $k$ unique $z$ values $\{z_1,...,z_k\}$. I.e., for the set $Z_k = \{z_1,...,z_k\}$, $I_k(f;z)$ denotes the mutual information between $f$ and $z$ in the distribution $p_k(f, z) = p(f|z) p_k(z)$, where $p(f|z)$ matches the corresponding conditional in $p(f,z)$, $p_k(z) \propto p(z)$ for $z_i \in Z_k$, and $p_k(z) = 0$ for $z_j \notin Z_k$.

Under some minor restrictions on how we sample possible $kz$-subsets, the corresponding expected mutual information $\expect_{Z_k} [I_k(f;z)]$ lower-bounds the mutual information $I(f;z)$ in the full joint $p(f,z)$. We can form a variational lower bound on $I_k(f;z)$ for the $kz$-subset based on $Z_k = \{z_1,...,z_k\}$ by sampling a $z_i$ from $Z_k$ in proportion to its marginal probability $p(z_i)$ in the full joint, then sampling a partial function $\hat{f}$ given $z_i$, and then measuring categorical cross entropy of a prediction $q(z_i|\hat{f})$ which is normalized over $z_j \in Z_k$. This categorical cross entropy plays the same role as the cross entropy in Eqn.~\ref{eq:variational_ent_bound}, and lets us place a lower bound on the restricted mutual information $I_k(f;z)$. By sampling $Z_k$ appropriately from the set of all $k$-ary subsets of the latent space for the full joint distribution, and then maximizing $I_k(f; z)$, we can maximize an expectation over lower bounds that provides a lower bound on the full mutual information $I(f;z)$.

The prediction $q(z_i|\hat{f})$ can be made by learning functions $\phi_f(\hat{f})$ and $\phi_z(z)$ which respectively embed partially observed functions and latent variables into a shared, high-dimensional space. To compute $q_k(z_i|\hat{f})$ for a given $\hat{f}$ and set $Z_k = \{z_1,...,z_k\}$, we can compute the dot product $\phi_f(\hat{f})^{\top} \phi_z(z_i)$ for each $z_i \in Z_k$ and normalize by softmax. Given a sufficiently high-dimensional shared embedding space and sufficiently powerful $\phi_f$/$\phi_z$, this approach to representing $q_k(z_i|\hat{f})$ is sufficient for representing (almost) any set of desired conditionals over $\hat{f}$ and $z_i$. This can be argued by reference to Reproducing Kernel Hilbert Spaces.

We can compute a bound on $I(f;z)$ using $kz$-subsets and softmaxed dot products between function and latent variable embeddings quite efficiently, even for large $k$. One potential benefit of this approach is the freedom to choose arbitrary architectures for the embedding functions $\phi_f$/$\phi_z$, in contrast to the more obvious variational bound, which requires a normalized density estimator for $q(z|f)$.




\end{document}